\documentclass[11pt]{article}
\usepackage[margin=1in]{geometry}

\usepackage{amsmath,amsthm,amssymb,amsfonts,cancel}
\usepackage{thmtools,enumitem,subfigure}
\usepackage{algcompatible}
\usepackage{algorithm}
\usepackage[noend]{algpseudocode}
\usepackage{mathtools}
\usepackage{dsfont}
\usepackage{tcolorbox}
\usepackage{bbm}
\usepackage{todonotes}
\usepackage{tensor}
\usepackage{caption}
\usepackage{bm}
\usepackage{caption}
\usepackage{hyperref}
\usepackage{placeins}
\usepackage{authblk}
\usepackage{epsfig}
\usepackage{natbib}
\usepackage[utf8]{inputenc}
\usepackage[T1]{fontenc}
\usepackage[stable]{footmisc}
\usepackage{multirow}

\renewcommand{\S}{\mathcal{S}}
\renewcommand{\P}{\mathcal{P}}
\newcommand{\D}{\mathcal{D}}
\newcommand{\A}{\mathcal{A}}

\newcommand{\dom}[1]{\texttt{dom}\,\left( #1 \right)}
\newcommand{\Ind}[1]{\mathds{1}_{\left[ #1 \right]}}

\newcommand{\Exp}[1]{\mathbb E \left[ #1 \right]} 

\renewcommand{\Pr}{\mathbb{P}}
\newcommand{\pfail}{\delta}
\newcommand{\Qhat}[2]{\mathbf{Q}_{#1}^{#2}}
\newcommand{\Vhat}[2]{\mathbf{V}_{#1}^{#2}}

\mathchardef\mhyphen="2D 

\DeclareMathOperator*{\argmax}{\arg\!\max}
\DeclareMathOperator*{\dsmax}{\max}

\let\originalleft\left
\let\originalright\right
\renewcommand{\left}{\mathopen{}\mathclose\bgroup\originalleft}
\renewcommand{\right}{\aftergroup\egroup\originalright}
\newcommand\blfootnote[1]{%
  \begingroup
  \renewcommand\thefootnote{}\footnote{#1}%
  \addtocounter{footnote}{-1}%
  \endgroup
}

\newtheorem{theorem}{Theorem}
\numberwithin{theorem}{section}
\newtheorem{definition}[theorem]{Definition}

\makeatletter
\let\@fnsymbol\@arabic
\makeatother

\usepackage{textcomp}

\newcommand{\Bhat}{\mathbf{B}}

\newcommand{\taumin}{\tau_{\text{min}}}

\PassOptionsToPackage{normalem}{ulem}
\usepackage{ulem}

\providecolor{added}{rgb}{0,0,1}
\providecolor{deleted}{rgb}{1,0,0}

\begin{document}

\title{Single-partition adaptive Q-learning}
\author{João Pedro Araújo\thanks{Email: \texttt{joao.p.araujo@tecnico.ulisboa.pt}} \qquad Mário Figueiredo$^2$\qquad Miguel Ayala Botto$^3$}
\date{\small{$^1$Instituto Superior Técnico, Universidade de Lisboa \\
$^2$Instituto de Telecomunicações; Instituto Superior Técnico, Universidade de Lisboa, Portugal.\\
$^3$IDMEC, Instituto Superior Técnico, Universidade de Lisboa, Portugal.}}
	\maketitle

	\begin{abstract}
This paper introduces \textit{single-partition adaptive Q-learning} (SPAQL), an algorithm for model-free episodic reinforcement learning (RL), which adaptively partitions the state-action space of a \textit{Markov decision process} (MDP), while simultaneously learning a time-invariant policy (\textit{i. e.}, the mapping from states to actions does not depend explicitly on the episode time step) for maximizing the cumulative reward. The trade-off between exploration and exploitation is handled by using a mixture of \textit{upper confidence bounds} (UCB) and Boltzmann exploration during training, with a temperature parameter that is automatically tuned as training progresses. The algorithm is an improvement over \textit{adaptive Q-learning} (AQL). It converges faster to the optimal solution, while also using fewer arms. Tests on episodes with a large number of time steps show that SPAQL has no problems scaling, unlike AQL. Based on this empirical evidence, we claim that SPAQL may have a higher sample efficiency than AQL, thus being a relevant contribution to the field of efficient model-free RL methods. \blfootnote{\\ The code for the experiments is available at \url{https://github.com/jaraujo98/SinglePartitionAdaptiveQLearning}.}

\end{abstract}
\newpage
\setcounter{tocdepth}{2}
\tableofcontents
\newpage

\section{Introduction}

\textit{Reinforcement learning} (RL) is an area within machine learning that studies how agents (such as humanoid robots, self-driving cars, or computer programs that play chess) can learn how to perform their tasks without being explicitly told how to do so. The problem can be posed as that of learning a mapping from system states to agent actions. In order to do this, the agent observes and interacts with the system, choosing which action to perform given its current state. By choosing a certain action, the system transitions to a new state, and the agent may receive a reward or a penalty. Based on these rewards and penalties, the agent learns to assess the quality of actions. By setting the agent's objective to be reward maximization, it will learn to prefer good actions over bad ones. As an example, consider an agent being trained to play chess. The system is the chessboard, and the actions correspond to moving the pieces. Good actions lead to victory, while bad actions lead to defeat. The agent may receive rewards for capturing opponent pieces, and may receive penalties when having its own pieces captured. The sum of all rewards and penalties over a game is called the cumulative reward. Maximizing this cumulative reward is the objective of RL \citep{sutton2018}. This approach has led to many successful applications of RL in areas such as control of autonomous vehicles \citep{DLVehicles}, game playing \citep{Silver2018, openai2019dota}, healthcare \citep{Shortreed2011}, and education \citep{10.1145/3231644.3231672}, to name a few.

Existing RL methods can be divided into two main families: model-free and model-based. In model-based methods, the agent either has access to or learns a model of the system, which it uses to plan future actions. Although model-free methods are applicable to a wider array of problems, concerns with the empirical performance of model-free methods have been raised in the past due to their sample complexity \citep{pilco, trpo}. Two additional problems, besides sample complexity, are: the trade-off between exploration and exploitation, and dealing with \textit{Markov decision processes} (MDP) with continuous state and action spaces. Several solutions to deal with these two latter problems have been proposed. For example, exploration and exploitation can be balanced using a stochastic policy, such as $\varepsilon$-greedy \citep{sutton2018}. MDPs with continuous state and action spaces can be dealt with by using function approximators, such as neural networks.

Recently, some theoretical work has addressed the sample complexity of model-free methods. Informally, sample complexity can be defined as the number of samples that an algorithm requires in order to learn. Sample-efficient algorithms require fewer samples to learn, and thus are more desirable to develop and use. There are several approaches to sample-efficient RL \mbox{\citep{yu2018towards}}. However, until recently, it was not known whether model-free methods could be proved to be sample-efficient. This changed when \citet{JAZBJ18} proposed a Q-learning algorithm with \textit{upper confidence bound} (UCB) exploration for discrete tabular MDPs, and showed that it had a sample efficiency comparable to \textit{randomized least-squares value iteration} (RLSVI), a model-based method proposed by \citet{Osband16}. Later, \citet{Song19} extended that algorithm to continuous state-action spaces. Their algorithm, \textit{net-based Q-learning} (NBQL), requires discretizing the state-action space with a network of fixed step size. This creates a trade-off between memory requirements and algorithm performance. Coarser discretizations are easier to store, but the algorithm may not have sufficient resolution to achieve satisfactory performance. Finer discretizations yield better performance, but involve higher memory requirements. In order to address this trade-off, \citet{Sinclair19} proposed \textit{adaptive Q-learning} (AQL). AQL introduces adaptive discretization into NBQL, starting with a single ball covering the entire state-action space, and then adaptively discretizing it in a data-driven manner. This adaptivity ensures that the relevant parts of the state-action space are adequately partitioned, while keeping coarse discretizations in regions that are not so relevant. Shortly after, \citet{Touati20} proposed ZoomRL, a variant of AQL that keeps the regret bound, while not requiring access to a packing oracle. However, ZoomRL requires prior knowledge of the Lipschitz constant of the Q function, which may not be readily available.

The papers mentioned in the previous paragraph consider time-variant value functions and learn time-variant policies. This means that a state-action space partition needs to be kept for each time step. However, for many practical purposes, time-invariant policies are sufficient to solve the problem satisfactorily (albeit not optimally). This is particularly true in problems with time-invariant dynamics, such as the classical \texttt{Pendulum} problem. Although it seems straightforward to modify AQL to learn time-invariant policies, empirical tests have shown that doing this modification na\"ively leads to poor empirical performance. This paper introduces an improved version of AQL specifically tailored to learn time-invariant policies that performs as well as AQL empirically (and in some cases outperforms it). This paper's main objective is to provide an empirically-validated algorithm that may pave the road for future developments in sample-efficient model-free RL algorithms.

\subsection{Contributions}
\label{section:contributions}

This paper proposes and  empirically tests a new algorithm: \textit{single-partition adaptive Q-learning} (SPAQL). The proposed algorithm keeps a single estimate for the value and Q functions, which is used at every time step. Exploration is balanced with exploitation by using a mixture of UCB and Boltzmann exploration with a cyclic adaptive temperature schedule.

The proposed SPAQL algorithm is compared with the original AQL in the \textit{Oil Discovery} and \textit{Ambulance Routing} problems, on which AQL was originally tested \citep{Sinclair19}. A large number of tests with different random seeds are carried to ensure the statistical significance of the results. In some of the problems, SPAQL outperforms AQL, while at the same time converging faster and using a smaller number of arms (\textit{i.e.}, possible actions).

\subsection{Related Work}
\label{section:related}

\subsubsection{Efficient model-free RL} 
There has been a considerable amount of work on sample-efficient model-free RL algorithms. For a thorough list of references, the reader is referred to the introductory sections of the papers by \citet{Sinclair19} and \citet{Touati20}. More recently, \citet{neustroev2020generalized} introduced a unified framework for studying sample-efficient UCB-based algorithms in the tabular setting, which includes the previous algorithms as particular cases. With a few exceptions (notably the works of \citet{Sinclair19} and \citet{neustroev2020generalized}), this line of work has been mostly theoretical, with little to no empirical validation.

An efficient Q-learning algorithm with UCB in the infinite-horizon setting has been proposed by \citet{Wang2020Q-learning}. Their algorithm learns time-invariant policies, but only considers the tabular case, which means that it does not work with continuous state-action spaces, unlike the algorithm proposed in this paper.

\subsubsection{Boltzmann exploration} 
Previous works have used UCB to deal with exploration. However, early tests with SPAQL showed that it was not enough, and Boltzmann exploration was introduced. This exploration technique is a standard tool in RL \citep{cesa2017boltzmann}, and it has been seen performing better than other strategies, such as UCB, in selected problems \citep{kuleshov2014,Tijsma2016}.

Boltzmann exploration has its drawbacks, namely the existence of multiple fixed points \citep{10.5555/924039}. To deal with this issue, alternative softmax-based operators have been proposed by \citet{pmlr-v70-asadi17a} and \citet{DBLP:journals/corr/abs-1903-05926}. The approach followed in this paper uses the classical softmax function, while keeping track of the best policy found so far, therefore clipping the instability associated with Boltzmann exploration.

The main challenge involved with the use of Boltzmann exploration is setting the temperature schedule. A straightforward approach, used by \citet{Tijsma2016}, is to set a constant temperature, which can be found using grid search or any other tuning method. Another approach is to decrease (anneal) the temperature as some function of the training iteration \citep{DBLP:journals/corr/abs-1903-05926}. The approach herein followed is the opposite one: start with a low temperature, and gradually increase it if the performance is not improving. This allows a gradual shift from exploitation to exploration. Once the policy improves,  the temperature is reset to a low value, and the procedure is repeated. This is similar to cyclical annealing schedules, which have already been used in supervised learning applications \citep{DBLP:conf/wacv/Smith17,n.2018superconvergence,DBLP:conf/iclr/HuangLP0HW17,fu_cyclical_2019_naacl,mu2018parameter,DBLP:conf/iclr/LoshchilovH17}. In all these methods, the cycle length is controlled by at least one user-defined parameter. In the algorithm proposed in this paper, the cycle reset occurs automatically, when an increase in performance is detected. The user only has to provide two parameters: one parameter controlling the rate at which the temperature heads towards the maximum value; another that works similarly to a doubling period factor. It also makes more sense to talk about cold restarts instead of warm restarts, since at each restart the temperature is set to the minimum value. \citet{Yamaguchi11} use a combination of UCB and Boltzmann exploration as exploration strategy. Unlike the approach followed in this paper, they use a monotonically decreasing schedule for the temperature.

\subsection{Outline of the paper}
Section~\ref{section:preliminaries} recalls some background concepts in RL and metric spaces. The proposed SPAQL algorithm is described and explained in Section~\ref{section:algorithm}. The experimental setup used to evaluate the algorithm, along with the analysis of the results, can be found in Section~\ref{section:experiments}. Finally, conclusions are drawn in Section~\ref{section:conclusion}. The full experimental results are presented in Appendix~\ref{section:full_experiments}.

\section{Background}
\label{section:preliminaries}

For background in RL, the reader is referred to the classic book by \citet{sutton2018}. The same notation as \citet{Sinclair19} will be used. This section recalls some basic concepts of RL and metric spaces, which are relevant for the subsequent presentation.

\subsection{Markov decision processes}
Since the algorithm proposed in this paper builds upon AQL, the assumptions made by \citet{Sinclair19} are kept, namely that  MDPs have finite horizon. An MDP is a 5-tuple $(\S, \A, H, \Pr, r)$, where:
\begin{itemize}
    \item $\S$ denotes the set of \textit{environment states};
    \item $\A$ is the \textit{set of actions} of the agent interacting with the environment;
     \item  $H$ is the \textit{number of steps} in each episode (also called the \textit{horizon});
     \item $\Pr$ is the \textit{transition kernel}, which assigns to each triple $(x, a, x')$ the probability of reaching state $x' \in \S$, given that action $a \in \A$ was chosen while in state $x \in \S$; this is denoted as $x' \sim \Pr(\cdot \mid x,\ a)$ (unless otherwise stated, $x'$ represents the state to which the environment transitions when action $a$ is chosen while in state $x$ under transition kernel $\Pr$);
     \item $r: \S \times \A \rightarrow R \subseteq \mathbb{R}$ is the \textit{reward function}, which assigns a reward (or a cost) to each state-action pair (\citet{Sinclair19} use $R = [0, 1]$).
\end{itemize}

In this paper, the only limitation imposed on the state and action spaces is that they are bounded\footnote{It is easy to take an unbounded set and map it into a bounded one (for example, the set of real numbers can be mapped into the unit interval using the logistic function).}. Furthermore, it is assumed that the MDP has time-invariant dynamics, meaning that neither the transition kernel nor the reward function vary with time. There is previous work, such as that by \citet{NSMDP2019}, which uses the term 
``stationary'' when referring to time-invariant MDPs.

The system starts at the initial state $x_1$. At each time step $h \in \{1,..., H\}$ of the MDP, the agent receives an observation $x_{h} \in \S$, chooses an action $a_h \in \A$, receives a reward $r_h = r(x_{h},a_h)$, and transitions to state $x_{h+1} \sim \Pr(\cdot \mid x_{h},\ a_h)$. The objective of the agent is to maximize the cumulative reward $\sum_{h=1}^H r_h$ received throughout the $H$ steps of the MDP. This is achieved by learning a function $\pi : \S \rightarrow \A$ (called a \textit{policy}) that maps states to actions in a way that maximizes the accumulated rewards. This function is then used by the agent when interacting with the system. If the policy is independent of the time step, it is said to be time-invariant.

\subsection{Q-learning}

One possible approach for learning a good (eventually optimal) policy is \textit{Q-learning}. The idea is to associate with each state-action pair $(x,a)$ a number that expresses the \textbf{q}uality (hence the name Q-learning) of choosing action $a$ given that the agent is in state $x$. The value of a state $x$ at time step $h$ is defined as the expected cumulative reward that can be obtained from that state onward, under a given policy $\pi$:

\begin{equation}
V^\pi_h(x) := \Exp{\sum_{i=h}^H r(x_{i}, \pi(x_{i})) ~\Big|~ x_h = x}.
\end{equation}

This is equivalent to averaging all cumulative rewards that can be obtained under policy $\pi$ until the end of the MDP, given that at time step $h$ the system is in state $x$. By taking into account all rewards until the end of the MDP, the value function provides the agent with information regarding the rewards in the long run. This is relevant, since the state-action pairs yielding the highest rewards in the short run may not be those that yield the highest cumulative reward. However, if $H$ is large enough and there is a large number of state-action pairs, it may be impossible (or impractical) to compute the actual value of $V^\pi_h(x)$ for all $1 \leq h \leq H$ and $x \in \S$, and some approximation has to be used instead. For example, a possible estimator samples several paths at random, and averages the cumulative rewards obtained \citep{sutton2018}.

The \textit{value function} $V^\pi_h : \S \rightarrow \mathbb{R}$ provides information regarding which states are more desirable. However, it does not take into account the actions that the agent might choose in a given state. The \textit{Q function},

\begin{equation}
Q^\pi_h(x, a) := r(x,a) + \Exp{\sum_{i=h+1}^H r(x_{i}, \pi(x_{i})) ~\Big|~ x_h = x,\ a_h = a} = r(x,a) + \Exp{V^\pi_h(x')~\Big|~ x,\ a},
\end{equation}
takes this information into account since its domain is the set of all possible state and action pairs. This allows the agent to rank all possible actions at state $x$ according to the corresponding values of Q, and then make a decision regarding which action to choose. Notice that the second expectation in the previous equation is with respect to $x'  \sim \Pr(\cdot \mid x,\ a)$.

The functions $V^\star_h(x) = \sup_\pi V^\pi_h(x)$ and $Q^{\star}_h(x, a) = \sup_\pi Q^\pi_h(x, a)$ are called the \textit{optimal value function} and the \textit{optimal Q function}, respectively. The policies $\pi^\star_V$ and $\pi^\star_Q$ associated with $V^\star_h(x)$ and $Q^{\star}_h(x, a)$ are one and the same, $\pi^\star_V=\pi^\star_Q=\pi^\star$; this policy is called the \textit{optimal policy} \citep{sutton2018}. There may be more than one optimal policy, but they will always share the same optimal value and optimal Q function. These functions satisfy the so-called \textit{Bellman equation}

\begin{equation}
\label{eq:bellman}
    Q^{\star}_h(x, a) = r(x, a) + \mathbb{E}\left[V^{\star}_h(x') \mid x, a\right].
\end{equation}


Q-learning is an algorithm for computing estimates $\Qhat{h}{}$ and $\Vhat{h}{}$ of the Q function and the value function, respectively. The updates to the estimates at each time step are based on Equation~\ref{eq:bellman}, according to

\begin{equation}
\Qhat{h}{}(x, a) \leftarrow (1-\alpha)\Qhat{h}{}(x, a) + \alpha\left(r(x,a) + \Vhat{h}{}(x')\right),
\end{equation}
where $\alpha \in [0, 1]$ is the  \textit{learning rate}. This update rule reaches a fixed point (stops updating) when the Bellman equation is satisfied, meaning that the optimal functions were found. Actions are chosen greedily according to the $\argmax$ policy

\begin{equation}
\pi(x) = \argmax_a \Qhat{h}{}(x, a).
\end{equation}

The greediness of the $\argmax$ policy may lead the agent to become trapped in local optima. To escape from these local optima, stochastic policies such as $\varepsilon$-greedy or Boltzmann exploration can be used. The $\varepsilon$-greedy policy chooses the greedy action with probability $1-\varepsilon$ (where $\varepsilon$ is a small positive number), and picks any action uniformly at random with probability $\varepsilon$ \citep{sutton2018}. Boltzmann exploration transforms the $\Qhat{h}{}$ estimates into a probability distribution (using softmax), and then draws an action at random. It is parametrized by a temperature parameter $\tau$, such that, when $\tau \rightarrow 0$, the policy tends to $\argmax$, whereas, for $\tau \rightarrow +\infty$, the policy tends to one that picks an action uniformly at random from the set of all possible actions.

Finding and escaping from local optima is part of the exploration/exploitation trade-off. Another way to deal with this trade-off is to use \textit{upper confidence bounds} (UCB). Algorithms that use UCB add an extra term $b(x,a)$ to the update rule

\begin{equation}
\Qhat{h}{}(x, a) \leftarrow (1-\alpha)\Qhat{h}{}(x, a) + \alpha\left(r(x,a) + \Vhat{h}{}(x') + b(x,a)\right),
\end{equation}
which models the uncertainty of the Q function estimate. An intuitive way of explaining UCB is to say that the goal is to choose a value of $b(x,a)$ such that, with high probability, the actual value of $Q^{\star}_h(x, a)$ lies within the interval $[\Qhat{h}{}(x, a), \Qhat{h}{}(x, a)+b(x,a)]$. A very simple example is $b(x,a) = 1/n_{(x,a)}$, where $n_{(x,a)}$ is the number of times action $a$ was chosen while the agent was in state $x$. As training progresses and $n_{(x,a)}$ increases, the uncertainty associated with the estimate of $\Qhat{h}{}(x, a)$ decreases. This decrease affects the following training iterations since, eventually, actions that have been less explored will have higher $\Qhat{h}{}(x, a)$ due to the influence of term $b(x,a)$, and will be chosen by the greedy policy. This results in exploration.

In the finite-horizon case (finite $H$), the optimal policy usually depends on the time step \citep{busoniu2010reinforcement}. One of the causes for this dependency is the constraint $V_{H+1}^{\pi}(x)=0$ for all $x \in \mathcal{S}$ (as mentioned by \citet{Sinclair19} and \citet[Section 6.5]{sutton2018}), since it forces the value function to be time-variant. This paper works with time-invariant value and Q functions, and thus this constraint poses a challenge. This issue will be dealt with in Section~\ref{section:algorithm}.

\subsection{Metric spaces}
A metric space is a pair $(X, \D)$ where $X$ is a set and $\D: X \times X \rightarrow \mathbb{R}$ is a function (called the \textit{distance function}) satisfying the following properties:

\begin{equation}
\begin{array}{lll}
    (\forall x, y \in X)\ \D(x,y) = 0 \Leftrightarrow x = y, &&  \\ 
    (\forall x, y \in X)\ \D(x,y) = \D(y,x), && \\
    (\forall x, y, z \in X)\ \D(x,z) \leq \D(x,y) + \D(y,z), && \\
    (\forall x,\ y \in X)\ \D(x,y) \geq 0. &&
\end{array}
\end{equation}

A ball $B$ with center $x$ and radius $r$ is the set of all points in $X$ which are at a distance strictly lower than $r$ from $x$, $B(x,r) = \{b \in X : \D(x, b) < r\}$. The diameter of a ball is defined as $\text{diam}(B) = \sup_{x, y \in B} \D(x, y)$. The diameter of the entire space is denoted $d_{max} = \text{diam}(X)$. Next, the concepts of \textit{covering}, \textit{packing}, and \textit{net} are recalled.

\begin{definition}[\citet{Sinclair19}]
An \textit{$r$-covering} of $X$ is a collection of subsets of $X$ that covers $X$ (\textit{i.e.}, any element of $X$ belongs to the union of the collection of subsets) and such that each subset has diameter strictly less than $r$.
\end{definition}

\begin{definition}[\citet{Sinclair19}]
\label{definition:net}
{A set of points $\P \subset X$} is an \textit{$r$-packing} if the distance between any two points in $\P$ is at least $r$.  An $r$-\textit{net} of $X$ is an $r$-packing such that $X \subseteq \cup_{x \in \P} B(x, r)$.
\end{definition}

\section{Algorithm}
\label{section:algorithm}

\begin{algorithm}[t!]
    \caption{Auxiliary functions}
    \label{alg:aux}
    \begin{algorithmic} [1]
        \Procedure{Evaluate Agent}{$\P, \S, \A, H, N$}
	    \State Collect $N$ cumulative rewards from \textproc{Rollout}$(\P, \S, \A, H)$
	    \State Return the average cumulative reward
		\EndProcedure
		\Procedure{Rollout}{$\P, \S, \A, H$}
		\State Receive initial state $x_1$
		\For{each step $h \gets 1, \ldots, H$}
		    \State Pick the ball $B_{sel} \in \P$ by the selection rule $B_{sel} = \displaystyle \argmax_{B \in \texttt{RELEVANT}(x_h)} \Qhat{}{}(B)$
			\State Select action $a_h = a$ for some $(x_h, a) \in \dom{B_{sel}}$
			\State Play action $a_h$, record reward $r_h$ and transition to new state $x_{h+1}$
		\EndFor
		\State Return cumulative reward $\displaystyle \sum_h r_h$
		\EndProcedure
		\Procedure{Split Ball}{$B$}
		    \State Set $B_1, \ldots B_n$ to be an $\frac{1}{2}r(B)$-packing of $\dom{B}$, and add each ball to the partition $\P$ (see Definition~\ref{definition:net})
		    \State Initialize parameters $\Qhat{}{}(B_i)$ and $n(B_i)$ for each new ball $B_i$ to inherit values from the parent ball $B$
		 \EndProcedure
		 \Procedure{Boltzmann Sample}{$\Bhat$, $\tau$}
		    \State Normalize values in $\Bhat$ by dividing by $\max(\Bhat)$ (this helps to prevent overflows)
		    \State Sample a ball $B$ from $\Bhat$ by drawing a ball at random with probabilities following the distribution
		    \[
		    P(B = B_i) \sim \exp\left(\Qhat{}{}(B_{i})/\tau\right)
		    \]
		 \EndProcedure
	\end{algorithmic}
\end{algorithm}

\begin{algorithm*}[t!]
	\caption{Single-partition adaptive $Q$-learning}
	\label{alg}
	\begin{algorithmic}[1]
		\Procedure{Single-partition adaptive $Q$-learning}{$\S, \A, \D, H, K, N, \xi, \taumin, u, d$}
			\State Initialize partitions $\P$ and $\P'$ containing a single ball with radius $d_{max}$ and $\Qhat{}{} = H$
			\State Initialize $\tau$ to $\taumin$
			\State Calculate agent performance using \textproc{Evaluate Agent}($\P, \S, \A, H, N$)
			\For{each episode $k = 1, \ldots, K$}
				\State Receive initial state $x_1^k$
				\For{each step $h = 1, \ldots, H$}
				    \State Get a list $\Bhat$ with all the balls $B \in \P'$ which contain $x_h^k$
					\State Sample the ball $B_{sel}$ using \textproc{Boltzmann Sample}$(\Bhat, \tau)$
					\State Select action $a_h^k = a$ for some $(x_h^k, a) \in \dom{B_{sel}}$
					\State Play action $a_h^k$, receive reward $r_h^k$ and transition to new state $x_{h+1}^k$
					\State Update Parameters: $v = n(B_{sel}) \gets n(B_{sel}) + 1$ \\ 
					\State $\Qhat{}{k+1}(B_{sel}) \gets (1 - \alpha_v)\Qhat{}{k}(B_{sel}) + \alpha_v\left(r_h^k + \Vhat{}{k}(x_{h+1}^k) + b_{\xi}(v)\right)$ where \\
					\State \indent $\Vhat{}{k}(x_{h+1}^k) = \min(H, \displaystyle \dsmax_{B \in  \texttt{RELEVANT}(x_{h+1}^k)} \Qhat{}{k}(B))$ (see Section~\ref{section:algorithm})
					\If{$n(B_{sel}) \geq \left( \frac{d_{max}}{r(B_{sel})}\right)^2$}
					  \textproc{Split Ball}$(B_{sel})$
					\EndIf
				\EndFor
				\State Evaluate the agent using \textproc{Evaluate Agent}($\P', \S, \A, H, N$)
				\If{agent performance improved}
				    \State Copy $\P'$ to $\P$ (keep the best agent)
				    \State Reset $\tau$ to $\taumin$
				    \State Decrease $u$ using some function of $d$ (for example, $u \gets u^d$, assuming $d < 1$)
				\Else
				    \State Increase $\tau$ using some function of $u$ (for example, $\tau \gets u\tau$)
				    \If{more than two splits occurred}
				        \State Copy $\P$ to $\P'$ (reset the agent)
				        \State Reset $\tau$ to $\taumin$
				    \EndIf
				\EndIf
			\EndFor
		\EndProcedure
    \end{algorithmic}
\end{algorithm*}

The proposed SPAQL algorithm builds upon AQL \citep{Sinclair19}. The change proposed aims at tailoring the algorithm to learn time-invariant policies. The main difference is that only one state-action space partition is kept, instead of one per time step; \textit{i.e.}, $\Qhat{h}{k} := \Qhat{}{k}$ and $\Vhat{h}{k} := \Vhat{}{k}$, where $k$ denotes the current training iteration. The superscript $k$ is used to distinguish between the estimates being used in the update rules (denoted $\Qhat{}{k}$ and $\Vhat{}{k}$) and the updated estimates (denoted $\Qhat{}{k+1}$ and $\Vhat{}{k+1}$). In order to simplify the notation, the superscript may be dropped when referring to the current estimate (denoted $\Qhat{}{}$ and $\Vhat{}{}$).

\subsection{Auxiliary functions}

Before describing the algorithm, it is necessary to define some auxiliary functions (Algorithm~\ref{alg:aux}). A measure of an agent's performance is obtained by performing a full episode under the current policy, and recording the cumulative reward. This is called a rollout, and is performed using function \textproc{Rollout}. The arguments that are given as input to this function are the ones describing the environment (state and action spaces $\S$ and $\A$, and horizon $H$) and the agent (partition $\P$, with estimates $\Qhat{}{}$). Since there might be random elements in the environment (stochastic transitions, for example) and in the agent (stochastic policies), it is necessary to estimate the average performance of the agent by doing several rollouts and averaging the cumulative rewards. This estimation is done by function \textproc{Evaluate Agent}. Besides the input arguments required for the \textproc{Rollout} function, \textproc{Evaluate Agent} also requires the number of rollouts $N$ to perform.

Functions \textproc{Boltzmann Sample} and \textproc{Split Ball} are used during training for action selection and ball splitting, respectively. Function \textproc{Boltzmann Sample} implements Boltzmann exploration. It has as arguments a list $\Bhat$ of balls which contain the given state $x \in \S$, and a temperature parameter $\tau$. It then draws a ball at random from $\Bhat$ according to the distribution induced by the values of $\Qhat{}{}$ and temperature $\tau$. Function \textproc{Split Ball} receives as input a ball $B$ and covers it with smaller balls of radius $r(B)/2$, where $r(B)$ is the radius of the ball being split.

\subsection{Main algorithm}

The algorithm (Algorithm~\ref{alg}) keeps two copies of the state-action space partition. One ($\P$) is used to store the best performing agent found so far (performance is defined as the average cumulative reward obtained by the agent). The other copy of the partition ($\P'$) is modified during training. At the end of each training iteration, the performance of the agent with partition $\P'$ is evaluated. If it is better than the performance of the previous best agent (with partition $\P$), the algorithm keeps the new partition ($\P \gets \P'$) and continues training. However, if performance decreases, then, at any moment, it is able to restart the agent and retrain from the previously found best one. In this way, the algorithm ensures that at the end of training the best agent which was found is returned. Another advantage of keeping only the partition associated with the best performance is that it forces an increase in the number of arms to correspond to an improvement in performance, thus preventing over-partitioning of the state-action space. At first, both partitions contain a single ball $B$ with radius $d_{max}$ (which ensures it covers the entire state-action space). The value of $\Qhat{}{}(B)$ is optimistically initialized to $H$, the episode length.

Each training iteration is divided into two parts. In the first one, a full episode (consisting of $H$ time steps) is played. The values of $\Qhat{}{}$ are updated in each time step, and splitting occurs every time the criterion is met. At the end of the episode, the agent is evaluated over $N$ runs, and the average cumulative reward computed. The second part of the training iteration modifies the policy according to the performance of the agent after training for an episode. To balance exploration with exploitation, a Boltzmann exploration scheme with an adaptive schedule is used. A temperature parameter $\tau$ allows  varying the policy between a deterministic $\argmax$ policy (for $\tau \rightarrow 0$) and a purely random one (for $\tau \rightarrow +\infty$). If the agent currently being trained achieved a better performance than previous agents, it may be somewhere worth exploiting. The value of $\tau$ is reset to a user-defined $\taumin\ (\approx 0)$ in order to ensure that the policy becomes greedy ($\argmax$). If the agent performs worse than the best agent, it may be because it is stuck in a local optimum, or simply in an uninteresting region. The value of $\tau$ is thus increased to make the policy behave in a more exploratory way. Empirical tests show that increasing $\tau$ by multiplying it by a factor $u > 1$ works well, although other schemes may also yield good results. The same tests also show that it is recommendable to normalize the values of $\Qhat{}{}$ before generating the probability distribution (see the definition of function \textproc{Boltzmann Sample} in Algorithm~\ref{alg:aux}), and vary $\tau$ between a minimum of $\taumin = 0.01$ (greedy) and a maximum of $10$ (random), at which updates to $\tau$ saturate.

Updates to $\Qhat{}{}$ are done according to 
\begin{equation}\label{eqn:qval_update}
\Qhat{}{k+1}(B_{sel}) \gets (1 - \alpha_v)\Qhat{}{k}(B_{sel}) + \alpha_v\left(r_h^k + \Vhat{}{k}(x_{h+1}^k) + b_{\xi}(v)\right),
\end{equation}
where $r_h^k$ is the reward obtained during training iteration $k$ on time step $h$, $\alpha_v$ is the learning rate, $\Vhat{}{k}(x_{h+1}^k)$ is the current estimate of the value of the future state, $b_{\xi}(v)$ is the bonus term related with the upper confidence bound (confidence radius), and $v$ is the number of times ball $B_{sel}$ has been \textbf{v}isited.

The learning rate is set as done by \citet{Sinclair19}, \textit{i.e.}, according to
\begin{equation}\label{eqn:learning_rate}
\alpha_v = \frac{H+1}{H+v}.
\end{equation}

Before defining $\Vhat{}{k}(x_{h+1}^k)$, it is necessary to recall the definition of domain of a ball and the definition of the set $\texttt{RELEVANT}(x)$ from the work of \citet{Sinclair19}. The \textit{domain} of a ball $B$ from a partition $\P$ is a subset of $B$ which excludes all balls $B' \in \P$ of a strictly smaller radius
\begin{equation}
\dom{B} = B \setminus \left( \bigcup_{B' \in \P : r(B') < r(B)} B'\right).
\end{equation}
In other words, the domain of a ball $B$ is the set of all points $b \in B$ which are not contained inside any other ball of strictly smaller radius than $r(B)$. A ball $B$ is said to be \textit{relevant} for a point $x \in \S$ if $(x, a) \in \dom{B}$, for some $a \in \A$. The set of all relevant balls for a state $x$ is denoted by $\texttt{RELEVANT}(x)$.
The definition of $\Vhat{}{k}(x_{h+1}^k)$ uses the expression proposed by \citet{Sinclair19},

\begin{equation}\label{eqn:future_value}
\Vhat{}{k}(x_{h+1}^k) = \min(H, \displaystyle \dsmax_{B \in  \texttt{RELEVANT}(x_{h+1}^k)} \Qhat{}{k}(B)),
\end{equation}
with the difference that it also holds for the final state, while \citet{Sinclair19} set $\Vhat{}{k}(x_{H+1}^k) = 0$, for all $x$. This boundary condition generates a dependency of the value function on the time step, that should be avoided when learning a time-invariant policy.

Finally, the term $b_{\xi}(v)$ is defined as

\begin{equation}\label{eqn:conf_radius_practical}
b_{\xi}(v) = \frac{\xi}{\sqrt{v}},
\end{equation}
where $\xi$ is called the \textit{scaling parameter} of the upper confidence bounds, and is defined by the user. This scaling parameter is defined by \citet{Sinclair19} as a function of problem constants,

\begin{equation}\label{eqn:conf_radius}
\xi = 2 \sqrt{H^3 \log\left(\frac{4HK}{\pfail}\right)} + 4Ld_{max}.
\end{equation}
where $\pfail \in [0,1]$ is a value related to the high-probability regret bound of the original algorithm, $L$ is the Lipschitz constant of the $Q^\star$ function, and $d_{max}$ is the radius of the initial ball containing the entire state-action space at the beginning of training.

Summarizing, the SPAQL algorithm includes the three rules used by the original AQL algorithm:
\begin{itemize}
	\item \textbf{Selection rule}: select a relevant ball $B$ for $x_h^k$ by drawing a sample from a distribution induced by the values of $\Qhat{}{}(B)$ and $\tau$ (breaking ties arbitrarily).  Select any action $a_h^k$ to play such that $(x_h^k, a_h^k) \in \dom{B}$.
	\item \textbf{Update parameters}: as done by \citet{Sinclair19} (increment $n(B)$, update $\Qhat{}{}(B)$ according to Equation~\ref{eqn:qval_update}), with the difference that the same $\Qhat{}{}$ function is considered at all time steps.
	\item \textbf{Re-partition the space}: if $n(B) \geq \left(d_{max}/r(B)\right)^2$, cover $\dom{B}$ with a $\frac{1}{2}r(B)$-Net of $\dom{B}$. Each new ball $B_i$ in the net inherits the $\Qhat{}{}(B_i)$ and $n(B_i)$ values from its parent ball $B$. This rule is kept equal to the original one proposed by \citet{Sinclair19}.
\end{itemize}

Two new rules are introduced in SPAQL to balance exploration and exploitation:
\begin{itemize}
	\item \textbf{Adapt the temperature}: if the agent's performance did not improve, the temperature $\tau$ is increased. If the performance improved, the policy is reset to greedy. As training progresses, and more visits are required to split existing balls, the temperature increase rate factor $u$ is decreased. This keeps the policy greedy for a longer time, and allows more splits to occur.
	\item \textbf{Reset the agent}: as training progresses, existing balls will be split. If a ball contains an optimum, further splits of that ball are required for the agent to get closer to it. If the agent performance has not increased after two splits (either in the same ball or on two different balls), then the agent and the policy are reset ($\P' \gets \P$, and $\tau$ is set to the minimum).
\end{itemize}

\section{Experiments}
\label{section:experiments}

In this section SPQAL is compared to AQL in the Oil Discovery and Ambulance Relocation problems, used by \citet{Sinclair19}. Their definition is recalled for convenience of the reader. The state and action space in both problems is $\S = \A = [0,1]$. The metric used is $\D((x,a), (x',a')) = \max\{|x - x'|, |a-a'|\}$. \blfootnote{The code associated with all experiments presented in this paper is available at \url{https://github.com/jaraujo98/SinglePartitionAdaptiveQLearning}}

\subsection{Experiment implementation}

For SPAQL agents, the implementation of the algorithm uses the tree data structure used by \citet{Sinclair19} to implement partition $\P_h^k$, but only keeps one tree per agent, instead of one per time step. The initial ball has center $(0.5, 0.5)$ and radius\footnote{Although the algorithm specifies that the initial radius should be $d_{max}=1$, this would end up covering a lot of space outside the state-action space. Experiments show that initializing the radius to the tighter $0.5$ value results in faster learning.} $0.5$ (a square of side $1$ centered at $(0.5, 0.5)$), allowing it to cover tightly the entire state-action space. Each split divides each ball into four equal balls, each with half of the radius of the parent ball (\textit{i. e.}, each square is split into four equal smaller squares).

Ball selection is done using a recursive algorithm. Since AQL uses $\argmax$ as policy, it only needs to check whether the state is contained in the ball, while keeping track of the ball which has the highest $\Qhat{}{}$ value. SPAQL uses a stochastic policy, and therefore it needs to check if the state-action pair is contained in the ball.

The temperature parameter $\tau$ schedule is controlled by parameters $u$ and $d$. The temperature is increased by multiplying by $u$ (a real number greater than $1$), and clipped at $10$ to avoid numerical problems. When a better agent is found, $u$ is updated to be $u^d$ (where $d$ is slightly smaller than $1$). In this way, more iterations are required before the policy becomes random, allowing for more exploitation and splitting.

\subsection{Experimental procedure and parameters}

\label{sec:exp_proc_param}

The parameters that are common to both algorithms are the number of training iterations (episodes) $K$, the episode length $H$, and the scaling value of the confidence bounds $\xi$. For each environment, the number of agents trained (to estimate average cumulative rewards) was the one found in the code provided with \citep{Sinclair19} ($25$ agents in the oil problem, and $50$ in the ambulance one).

The SPAQL algorithm has two additional parameters, $u$ and $d$, which control the temperature schedule. These were set to be $2$ and $0.8$, respectively, since they are seen to be acceptable over a wide range of experiments. The value of $\taumin$ was set to $0.01$. For a value of $u=2$, this means that $10$ iterations are enough to turn this policy into a random one\footnote{Since $0.01\times2^{10}\approx10$, value at which the temperature saturates.}.

In order to compare both algorithms, the impact of the scaling parameter $\xi$ is studied. A fixed set of $13$ scaling values

\begin{equation}
\xi \in \{0.01, 0.1, 0.25, 0.5, 0.75, 1, 1.25, 1.5, 1.75, 2, 3, 4, 5\},
\end{equation}
is used for both scenarios. Table \ref{tab:scaling_exp_parameters} contains the parameters used in the experiments whose results are reported in Figures~\ref{fig:oil_scaling_results} and~\ref{fig:ambulance_scaling_results}. The average cumulative reward at the end of training was calculated over the several agents for each algorithm, along with the $95\%$ confidence intervals. The trials that resulted in higher average cumulative rewards are used for the remaining analysis.

\begin{table}[ht]
\centering
\begin{tabular}{l|c}
\hline
\multicolumn{1}{c}{Parameter} & Value \\\hline
Episode length $H$               & 5     \\
Number of Oil episodes (training iterations)    $K$        & 5000  \\
Number of Oil agents              & 25   \\
Number of Ambulance episodes (training iterations) $K$           & 2000  \\
Number of Ambulance agents              & 50   \\
Number of evaluation Rollouts $N$             & 20   \\\hline
\end{tabular}
\caption{Parameters for the scaling experiments}
\label{tab:scaling_exp_parameters}
\end{table}

\subsection{Oil discovery}

\subsubsection{Experimental setup}
\label{section:experiments_set_up_oil}

In this problem, described by \citet{Mason2011}, an agent surveys a $1$D map in search of hidden ``oil deposits''. It is similar to a function maximization problem, where a cost is incurred every time a new estimate is made. This cost is proportional to the distance travelled. The state space is the set of locations that the agent has access to ($\S = [0,1]$). Since each action corresponds to the position of the agent in the next time step, the state space will match the action space ($\A = [0,1]$). The transition kernel is $\Pr_h(x' \mid x, a) = \Ind{x'=a}$, where $\Ind{A}$ is the indicator function (evaluates to $1$ if condition $A$ is true, and to $0$ otherwise). The reward function is $r_h(x,a) = \max \{0, f(a) - |x - a|\}$, where $f(a) \in [0, 1]$ is called the \textit{survey function}. This survey function encodes the location of the deposits ($f(a) = 1$ means that the exact location has been found). The same survey functions considered by \citet{Sinclair19} are considered in this paper:
\begin{itemize}
    \item Laplace survey function, $f(x) = e^{-\lambda |x - c|}$, with $\lambda \in \{1, 10, 50\}$;
    \item quadratic survey function, $f(x) = 1 - \lambda (x-c)^2$, with $\lambda \in \{1, 10, 50\}$.
\end{itemize}
The deposit is placed at approximately $c \approx 0.75$ (the actual location is $0.7 + \pi/60$). It is clear that the optimal policy is to choose the location of the deposit at every time step. Following this policy, an initial penalty is paid when moving from $0$ to $c$, but in the remaining steps the maximum reward is always obtained. Using this optimal policy, the cumulative reward can be bounded by

\[
1 - |0 - c| + (H-1) \times 1 = H - c.
\]
Applying this formula to the case under study (with $H=5$ and $c\approx 0.75$), the maximum reward is found to be approximately $ 4.25$.

\subsubsection{Experimental results}
\label{oil_experimental_results}

\begin{figure*}[t!]
\centering
\includegraphics[width=\columnwidth]{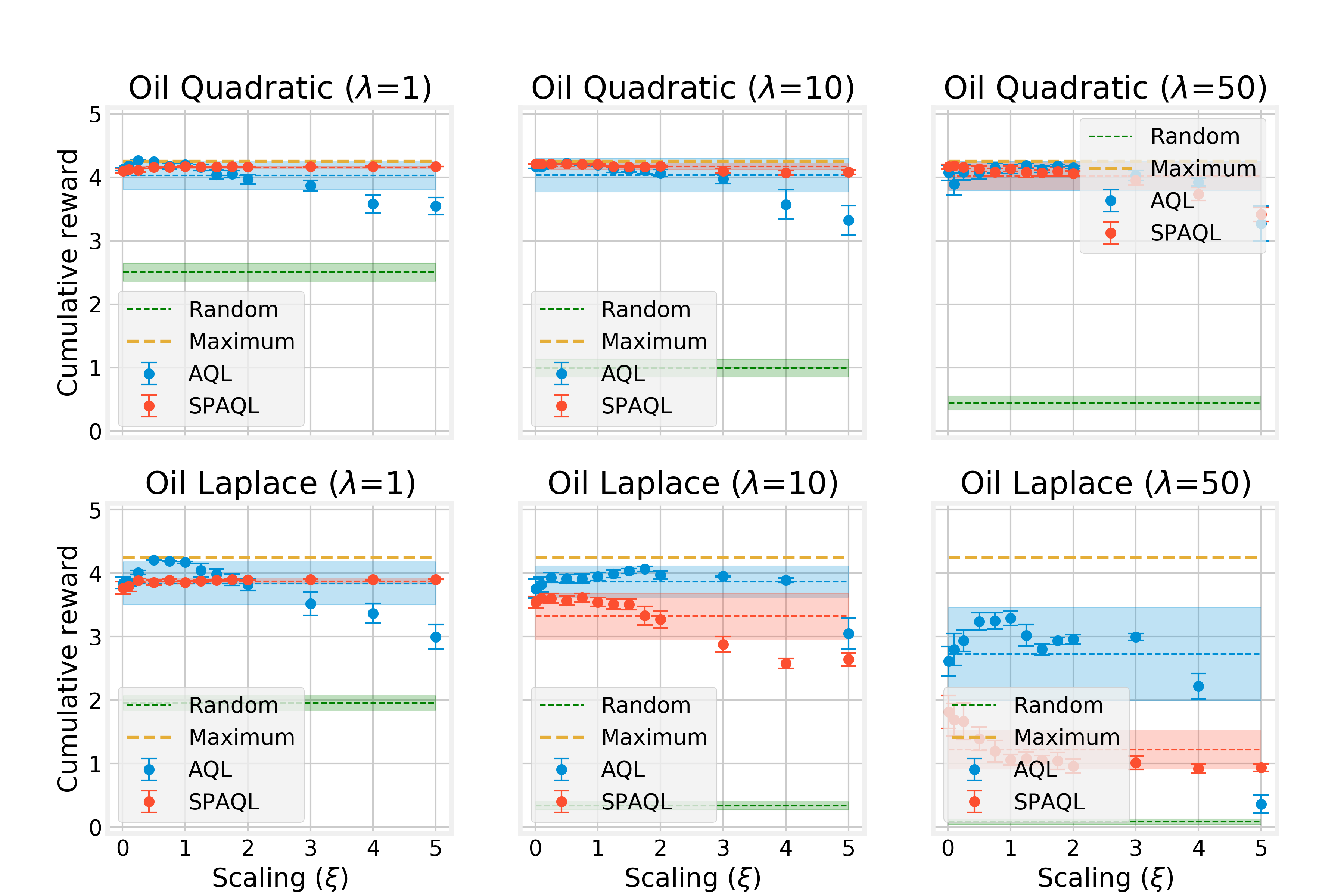}
\caption{
Comparison of the effect of different scaling parameter values on the average cumulative reward for the oil discovery problem with survey functions described in Section~\ref{section:experiments_set_up_oil}. Each dot corresponds to the average of the rewards obtained by the $25$ agents after $5000$ training iterations, and the error bars display the corresponding $95\%$ confidence interval. Dashed lines represent the average cumulative reward calculated over the $13$ scaling values listed in Section~\ref{sec:exp_proc_param}, and the shaded areas represent the corresponding standard deviation.
}
\label{fig:oil_scaling_results}
\end{figure*}

The effect of the scaling parameter $\xi$ on both AQL and SPAQL agents trained in the oil problem is shown in Figure~\ref{fig:oil_scaling_results}. 

The Laplace survey function results in more concentrated rewards. As the value of $\lambda$ is increased, the rewards become even more concentrated. This makes it harder for both types of agents to approach the maximum cumulative reward estimated previously ($4.25$). The scaling parameter affects more the AQL agents than the SPAQL ones (as seen by the size of the shaded regions). The AQL agents outperform the SPAQL agents in every instance of the problem with $\lambda \in [10, 50]$. When $\lambda=1$, half of the $\xi$ values result in higher rewards for the AQL agents, while the other half results in higher rewards for the SPAQL agents.

The quadratic survey function can also concentrate rewards, but does so in a more relaxed way than the Laplace survey function. This allows both types of agents to approach the maximum cumulative reward, even when $\lambda=50$. On average, SPAQL agents perform better than the AQL ones (higher average cumulative rewards and lower standard deviations). However, there always exists at least one scaling value for which the AQL agents achieve higher cumulative rewards than the SPAQL ones.

Two main conclusions that can be drawn from Figure~\ref{fig:oil_scaling_results}. The first one is that, for both types of agents, the value of $\xi$ should be picked from the interval $[0, H/3]$. The second one is that SPAQL agents are less sensitive to changes in $\xi$ than AQL agents. This is not surprising, since AQL agents rely solely on UCB for exploration, while SPAQL agents use UCB plus Boltzmann exploration.

\begin{figure*}[t!]
\centering
\includegraphics[trim={0 10cm 0 0},clip,width=\columnwidth]{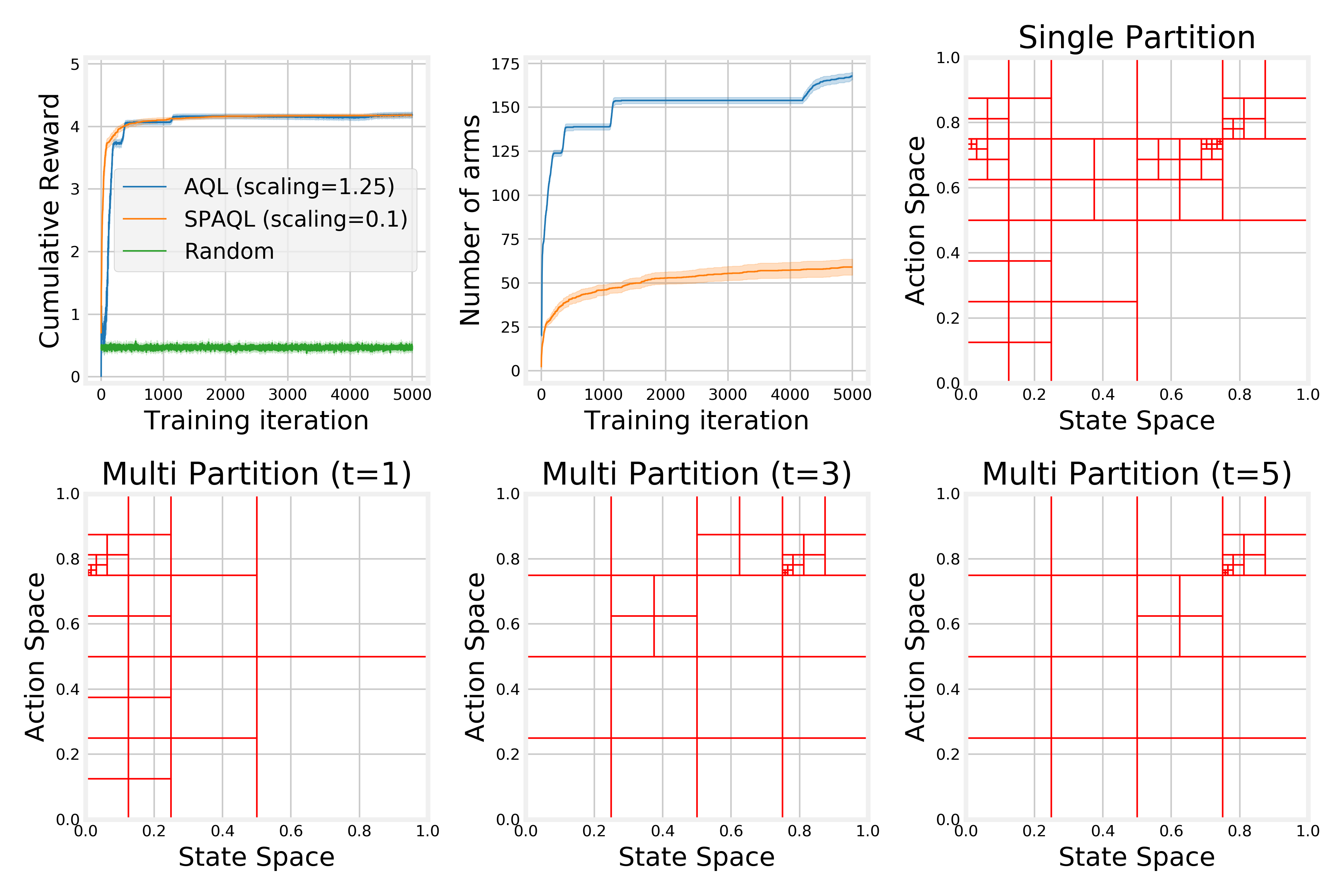}
\caption{Average cumulative rewards, number of arms, and best SPAQL agent partition for the agents trained in the oil problem with reward function $r(x, a) = \max\{0, 1 - 50(x-c)^2 - |x - a|\}$. Shaded areas around the solid lines represent the $95\%$ confidence interval.
}
\label{fig:oil_quadratic}
\end{figure*}

Shifting the focus from the scaling parameter to the learning process, Figure~\ref{fig:oil_quadratic} shows the average cumulative rewards obtained by the SPAQL and the AQL agents for the oil problem with quadratic survey function ($\lambda=50$), along with the partition learned by the best SPAQL agent. Looking at this partition, it can be seen that the neighborhoods of points $(0, 0.75)$ and $(0.75, 0.75)$, which correspond to the optimal policy, have been thoroughly partitioned, meaning that the agent located and exploited the location of the oil deposit. In the first training iterations, the SPAQL agents increase the cumulative rewards faster than the AQL agents. Both types of agents stabilize at the same level of cumulative reward, with the difference that the SPAQL agents use around one third of the arms of the AQL agents. The results of the other tests in the oil problem are presented in Table~\ref{table:results_oil}, and the respective figures are shown in Appendix~\ref{section:full_experiments}. The SPAQL agents always increase their cumulative rewards faster than the AQL agents during the initial training iterations. However, in most cases (especially for the Laplace survey function,\, Figures~\ref{fig:oil_laplace_1},~\ref{fig:oil_laplace_10}, and~\ref{fig:oil_laplace_50}), they are eventually surpassed by the AQL agents. Looking at the associated partitions, it is clear that this happens due to a lack of exploitation by the SPAQL agents, especially when the value of $\lambda$ is low (meaning that the rewards are high over a wide area). Tuning the values of $u$ and $d$ may be a way to increase the cumulative rewards in SPAQL. However, it is clear in every case that the SPAQL agents find the location of the oil deposit. The difference between AQL cumulative rewards and SPAQL cumulative rewards at the end of training is due to the coarseness of the SPAQL partition and the structure of the reward function, which may greatly amplify very small differences in actions. For example, using the Laplace survey function with $\lambda=50$, choosing action $c$ results in reward $1$, while choosing action $0.99c$ (an error of $1\%$) results in a reward of $\approx 0.61$, a decrease of almost $40\%$. With the quadratic survey function, the same relative error of $1\%$ in the action leads to a $0.5\%$ reduction in reward. This explains why SPAQL agents perform as well as AQL agents when using the quadratic survey function with $\lambda=50$, but only achieve half of the average cumulative reward when using the Laplace survey function. These relatively small differences in rewards, however, are compensated by the lower number of arms.

\begin{table}[t!]
\centering
\begin{tabular}{llccc}
                                                     &                             & \multicolumn{3}{c}{$\lambda$ (Quadratic)}                                                                 \\ \cline{3-5} 
                                                     &                             & 1                 & 10              & 50              \\ \hline
\multicolumn{1}{l|}{\multirow{3}{*}{\begin{tabular}[c]{@{}l@{}}Average cumulative\\ reward\end{tabular}}} & \multicolumn{1}{l|}{Random} & $2.50 \pm 0.06$   & $0.99\pm0.06$   & $0.44\pm0.04$   \\
\multicolumn{1}{l|}{}                                & \multicolumn{1}{l|}{AQL}    & $\mathbf{4.26 \pm 0.01}$   & $\mathbf{4.22\pm0.01}$   & $\mathbf{4.19\pm0.04}$   \\
\multicolumn{1}{l|}{}                                & \multicolumn{1}{l|}{SPAQL}  & $4.17 \pm 0.00$   & $4.21\pm0.00$   & $\mathbf{4.18\pm0.03}$   \\ \hline
\multicolumn{1}{l|}{\multirow{2}{*}{\begin{tabular}[c]{@{}l@{}}Average number\\ of arms\end{tabular}}} & \multicolumn{1}{l|}{AQL}    & $155.72 \pm 4.47$ & $140.60\pm2.86$ & $167.84\pm2.09$ \\
\multicolumn{1}{l|}{}                                & \multicolumn{1}{l|}{SPAQL}  & $42.04 \pm 1.90$  & $35.08\pm1.10$  & $59.08\pm4.52$  \\ \hline
\\
                                                     &                             & \multicolumn{3}{c}{$\lambda$ (Laplace)}                         \\ \cline{3-5} 
                                                     &                             & 1               & 10              & 50              \\ \hline
\multicolumn{1}{l|}{\multirow{3}{*}{\begin{tabular}[c]{@{}l@{}}Average cumulative\\ reward\end{tabular}}} & \multicolumn{1}{l|}{Random} & $1.95\pm0.05$   & $0.33\pm0.03$   & $0.08\pm0.02$  \\
\multicolumn{1}{l|}{}                                & \multicolumn{1}{l|}{AQL}    & $\mathbf{4.21\pm0.01}$   & $\mathbf{4.07\pm0.04}$   & $\mathbf{3.29\pm0.11}$   \\
\multicolumn{1}{l|}{}                                & \multicolumn{1}{l|}{SPAQL}  & $3.90\pm0.00$   & $3.61\pm0.07$   & $1.81\pm0.26$   \\ \hline
\multicolumn{1}{l|}{\multirow{2}{*}{\begin{tabular}[c]{@{}l@{}}Average number\\ of arms\end{tabular}}} & \multicolumn{1}{l|}{AQL}    & $158.36\pm2.83$ & $195.08\pm2.70$ & $357.08\pm7.12$ \\
\multicolumn{1}{l|}{}                                & \multicolumn{1}{l|}{SPAQL}  & $39.28\pm1.89$  & $67.12\pm4.89$  & $57.28\pm7.55$  \\ \hline
\end{tabular}
\caption{Average cumulative rewards and number of arms in the oil discovery problem. The best performing agent for each value of $\lambda$ is shown in bold. When the Welch t-test does not find a statistical difference between the two performances, both are shown in bold.}
\label{table:results_oil}
\end{table}

\subsection{Ambulance routing}

\subsubsection{Experimental setup}
\label{section:experiments_set_up_ambulance}

This problem, described by \citet{Brotcorne2003}, is a stochastic variant of the previous one. The agent controls an ambulance that, at every time step, has to travel to where it is being requested. The agent is also given the option to relocate after fulfilling the request, paying a cost to do so. \citet{Sinclair19} use a transition kernel defined by $\Pr_h(x' | x, a) \sim \mathcal{F}_h$, where $\mathcal{F}_h$ denotes the request distribution for time step $h$. The reward function is $r_h(x'|x,a) = 1 - [ c |x - a| + (1 - c)|x' - a| ]$, where $c\in[0,1]$ models the trade-offs between the cost of relocation and the cost of traveling to serve the request.

It is not mandatory that $\mathcal{F}_h$ varies with the time step. However, if that is so, then in principle a time-invariant policy would not be a good choice for solving the problem. In this paper only time-invariant scenarios ($\mathcal{F}_h := \mathcal{F}$) are considered. The experimental setups considered are
\begin{itemize}
    \item $\mathcal{F} = $~Beta$(5,2)$, for $c \in \{0, 0.25, 1\}$  (where Beta$(a,b)$ is the Beta probability distribution, modelling concentrated request distributions)
    \item $\mathcal{F} = $~Uniform$(0,1)$, for $c \in \{0, 0.25, 1\}$ (modelling disperse request distributions)
\end{itemize}

The optimal policy depends on the value of $c$. \citet{Sinclair19} suggest two heuristics for both extreme cases ($c \in \{0, 1\}$). The ``No Movement'' heuristic is optimal when $c = 1$. In this case, the cost paid is only the cost to relocate, and therefore if the agent does not relocate it does not incur on any cost. This policy corresponds to the line $x=a$ in the state-action space. The ``Mean''\footnote{The original authors probably meant ``Mean'' instead of ``Median''.} heuristic is optimal when $c = 0$. In this case, the cost paid is only the cost of traveling to meet a request. Therefore, the agent should relocate to where the next request is more likely to appear. The empirical mean $\hat{\mu}$ of distribution $\mathcal{F}$ is a good estimator of this location. This policy corresponds to the horizontal line $a=\hat{\mu}$ in the state-action space.

Intuitively, the optimal policies for the values of $c$ in between $0$ and $1$ will be a mix of these two optimal policies.

\subsubsection{Experimental results}
\label{ambulance_experimental_results}

\begin{figure*}[t!]
\centering
\includegraphics[width=\columnwidth]{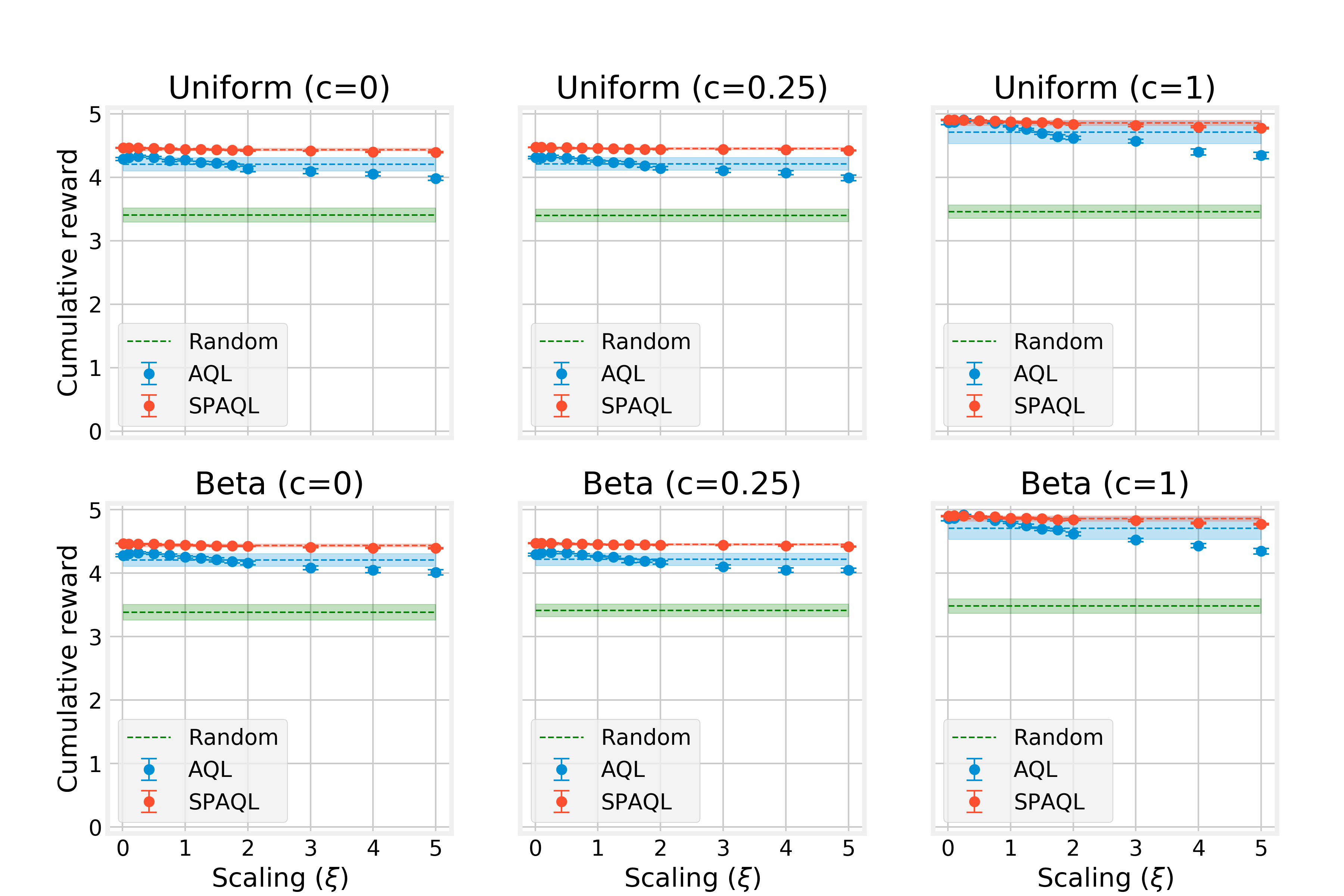}
\caption{Comparison of the effect different scaling parameter values on the average cumulative reward for the ambulance problem with arrival distributions described in Section~\ref{section:experiments_set_up_ambulance}. Each dot corresponds to the average of the rewards obtained by the $50$ agents after $2000$ training iterations, and the error bars display the corresponding $95\%$ confidence interval. Dashed lines represent the average cumulative reward calculated over the $13$ scaling values listed in Section~\ref{sec:exp_proc_param}, and the shaded areas represent the corresponding standard deviation.
}
\label{fig:ambulance_scaling_results}
\end{figure*}

The effect of the scaling parameter $\xi$ in the ambulance problem is seen in Figure~\ref{fig:ambulance_scaling_results}. Unlike the oil problem (Figure~\ref{fig:oil_scaling_results}), SPAQL agents perform better than AQL ones, independently of the value of $\xi$. The exception is the case $c=1$, where tuning of $\xi$ allows the AQL agents to match the cumulative rewards of SPAQL agents.
The average cumulative reward over the scaling values is always higher for the SPAQL agents, and the corresponding standard deviations are negligible when compared to those of AQL agents, meaning that in this problem the SPAQL agents have a very low sensitivity to the value of $\xi$.

\begin{figure*}[t!]
\centering
\includegraphics[trim={0 10cm 0 0},clip,width=\columnwidth]{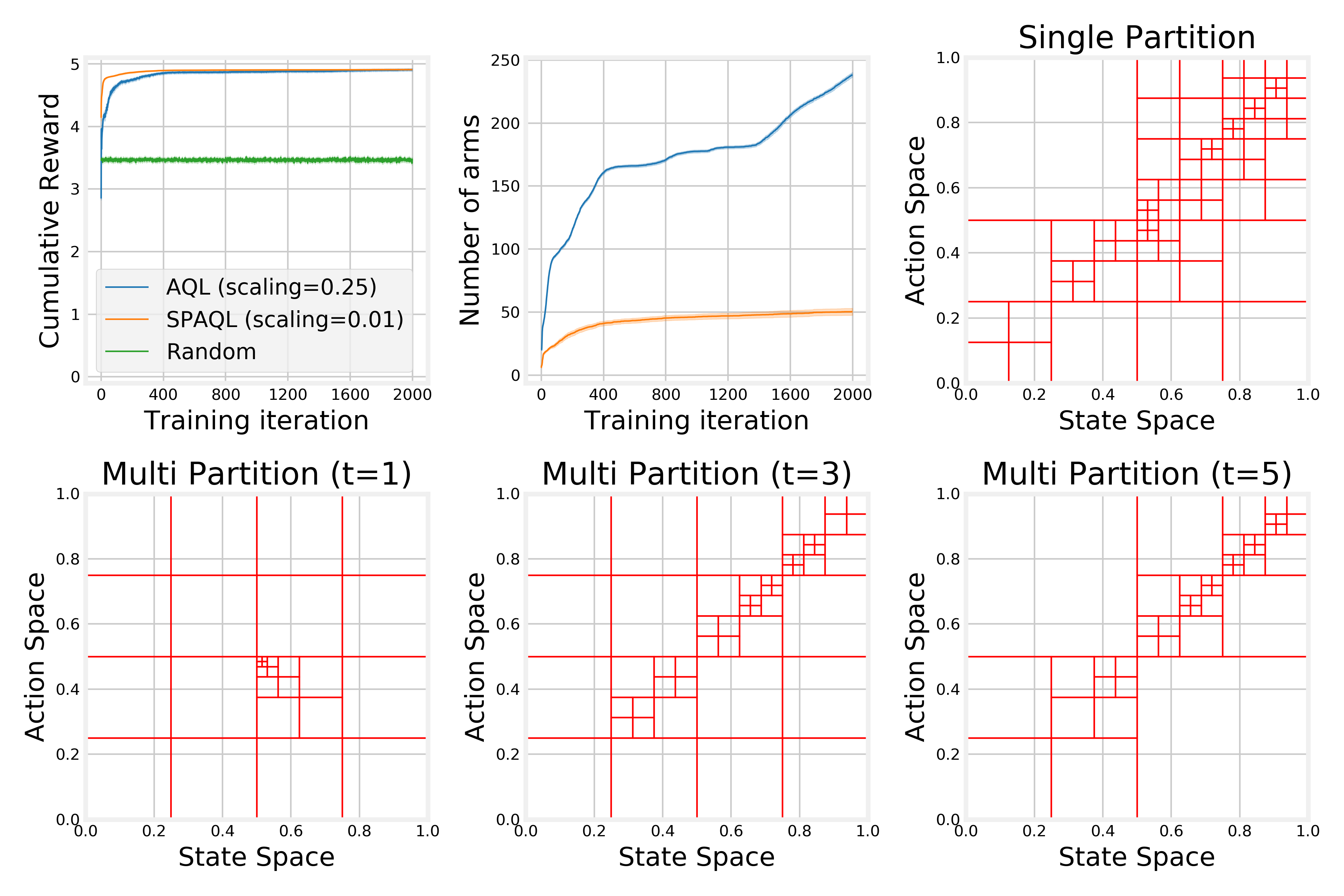}
\caption{Average cumulative rewards, number of arms, and best SPAQL agent partition for the agents trained in the ambulance problem with uniform arrival distribution and reward function $r(x, a) = \max\{0, 1 - |x - a|\}$. Shaded areas around the solid lines represent the $95\%$ confidence interval.
}
\label{fig:ambulance_uniform}
\end{figure*}

The full experimental results are presented in Table~\ref{table:results_ambulance}, and the respective images in Appendix~\ref{section:full_experiments}. Unlike the oil problem, in the ambulance problem the SPAQL agents are clearly and consistently better than the AQL ones, reaching higher cumulative rewards earlier in training, with much fewer arms (in some cases with only one fifth of the arms). This is consistent with the results in \citep{Tijsma2016}, who observed that Boltzmann exploration outperforms other exploration strategies in stochastic problems with rewards in the range $[0,1]$. Figure~\ref{fig:ambulance_uniform} shows the cumulative rewards for an ambulance problem with a uniform arrival distribution and $c=1$ (only relocation is penalized). Recall that the optimal policy is the ``No Movement'' one, which corresponds to the line $a = x$. This line appears finely partitioned, as would be expected. Keeping $c = 1$, but changing to a $\text{Beta}(5,2)$ distribution (Figure~\ref{fig:ambulance_beta_1}), it can be seen that the partition is now focused on the top right quadrant, where new arrivals are most likely to appear.

For this heuristic, the objective is to zero in on the diagonal as precisely as possible. As was already seen in the oil problem, AQL agents are better than SPAQL ones at this zeroing in. This is particularly seen in Figure~\ref{fig:ambulance_beta_1}, where the SPAQL partition is coarser than the AQL partitions in time steps $3$ and $5$. It is this difference that allows AQL agents to catch up to SPAQL agents. However, despite both types of agents stabilizing at the same cumulative reward level, SPAQL agents use much fewer arms, which gives them a competitive advantage over AQL agents.

For $c = 0$, the action under the ``Mean'' heuristic would be $a=0.5$ for the uniform arrival case, and $a\approx0.7$ for the $\text{Beta}(5,2)$ arrivals. These two policies correspond to horizontal lines, which could have been expected to be seen in the partitions in Figures~\ref{fig:ambulance_uniform_0} and~\ref{fig:ambulance_beta_0}, respectively. The neighborhoods of those lines are more partitioned than other areas of the state-action space, but not as much as the diagonal line in Figure~\ref{fig:ambulance_uniform}. Even though the partitions shown are coarser than the corresponding ones shown in \citep{Sinclair19}, the SPAQL agents reach higher rewards. This means that further partitions of the state-action space do not increase the average cumulative reward, and AQL agents may be needlessly over-partitioning the state-action space. This illustrates the remark made in Section~\ref{section:algorithm} regarding prevention of over-partitioning.

\begin{table}[!t]
\centering
\begin{tabular}{llccc}
                                                                                                          &                             & \multicolumn{3}{c}{$c$ (Uniform)}                          \\ \cline{3-5} 
                                                                                                          &                             & 0               & 0.25            & 1                \\ \hline
\multicolumn{1}{l|}{\multirow{3}{*}{\begin{tabular}[c]{@{}l@{}}Average cumulative\\ reward\end{tabular}}} & \multicolumn{1}{l|}{Random} & $3.41\pm0.03$   & $3.40\pm0.03$   & $3.46\pm0.03$    \\
\multicolumn{1}{l|}{}                                                                                     & \multicolumn{1}{l|}{AQL}    & $4.33\pm0.01$   & $4.33\pm0.02$   & $\mathbf{4.90\pm0.02}$    \\
\multicolumn{1}{l|}{}                                                                                     & \multicolumn{1}{l|}{SPAQL}  & $\mathbf{4.46\pm0.00}$   & $\mathbf{4.47\pm0.00}$   & $\mathbf{4.91\pm0.00}$    \\ \hline
\multicolumn{1}{l|}{\multirow{2}{*}{\begin{tabular}[c]{@{}l@{}}Average number\\ of arms\end{tabular}}}    & \multicolumn{1}{l|}{AQL}    & $244.10\pm2.51$ & $248.24\pm2.18$ & $238.40\pm1.84$  \\
\multicolumn{1}{l|}{}                                                                                     & \multicolumn{1}{l|}{SPAQL}  & $32.26\pm2.11$  & $29.44\pm2.03$  & $50.32\pm2.76$   \\ \hline
\\
                                                                                                          &                             & \multicolumn{3}{c}{$c$ (Beta)}                            \\ \cline{3-5} 
                                                                                                          &                             & 0               & 0.25           & 1                \\ \hline
\multicolumn{1}{l|}{\multirow{3}{*}{\begin{tabular}[c]{@{}l@{}}Average cumulative\\ reward\end{tabular}}} & \multicolumn{1}{l|}{Random} & $3.38\pm0.03$   & $3.41\pm0.03$  & $3.48\pm0.03$    \\
\multicolumn{1}{l|}{}                                                                                     & \multicolumn{1}{l|}{AQL}    & $4.32\pm0.02$   & $4.32\pm0.02$  & $\mathbf{4.92\pm0.01}$    \\
\multicolumn{1}{l|}{}                                                                                     & \multicolumn{1}{l|}{SPAQL}  & $\mathbf{4.47\pm0.01}$   & $\mathbf{4.47\pm0.00}$  & $4.91\pm0.00$    \\ \hline
\multicolumn{1}{l|}{\multirow{2}{*}{\begin{tabular}[c]{@{}l@{}}Average number\\ of arms\end{tabular}}}    & \multicolumn{1}{l|}{AQL}    & $244.04\pm3.14$ & $250.1\pm2.30$ & $239.54\pm1.95$  \\
\multicolumn{1}{l|}{}                                                                                     & \multicolumn{1}{l|}{SPAQL}  & $31.96\pm2.21$  & $29.56\pm1.97$ & $50.02\pm1.28$   \\ \hline
\end{tabular}
\caption{Average cumulative rewards and number of arms in the ambulance problem. The best performing agent for each value of $c$ is shown in bold. When the Welch t-test does not find a statistical difference between the two performances, both are shown in bold.}
\label{table:results_ambulance}
\end{table}

\subsubsection{Increasing the episode length}

\begin{figure*}[t!]
\centering
\includegraphics[width=\columnwidth]{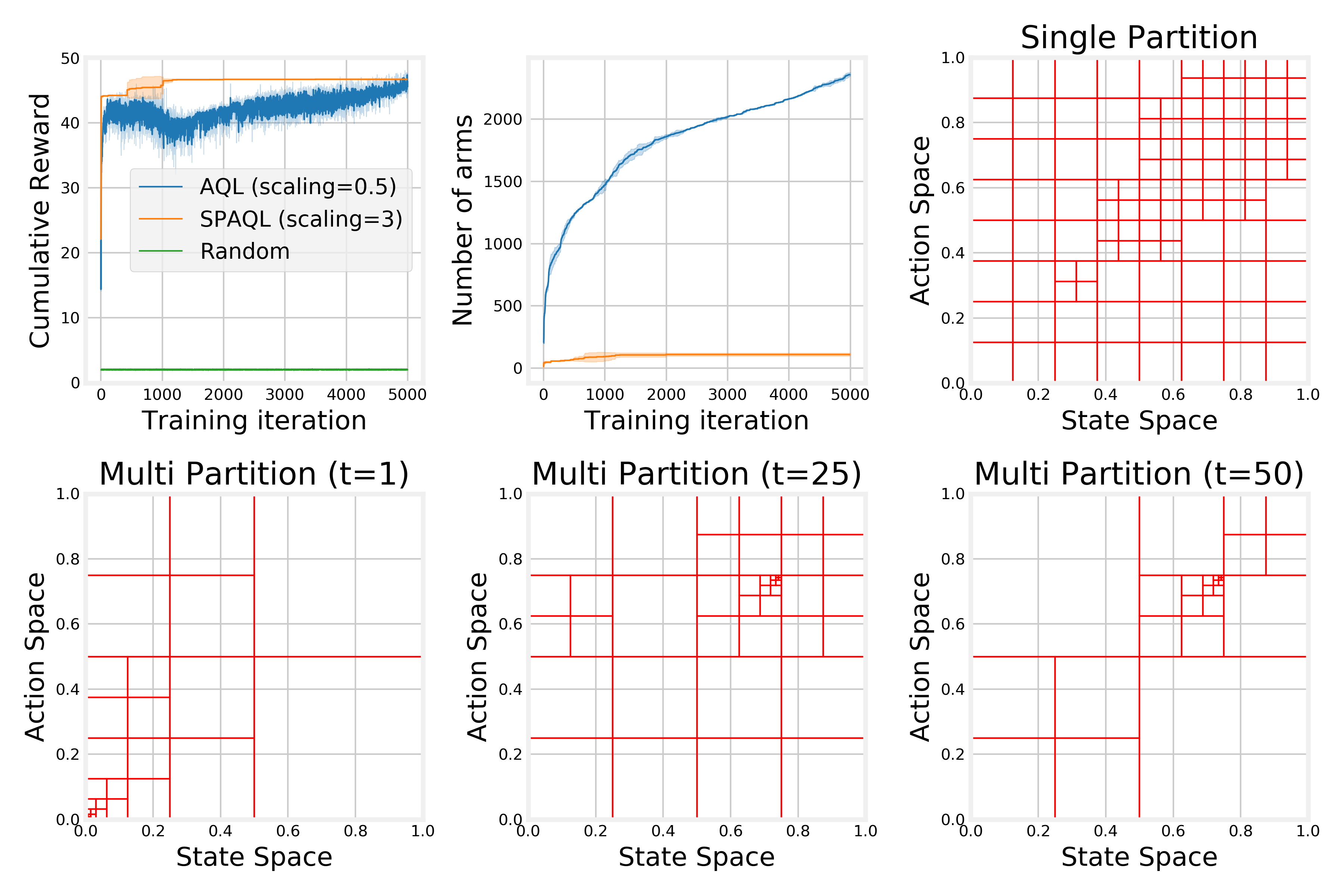}
\caption{Average cumulative rewards, number of arms, and partitions for the best SPAQL and AQL agents trained in the oil problem with Laplace survey function ($\lambda=1$). Shaded areas around the solid lines represent the $95\%$ confidence interval.
}
\label{fig:oil_laplace_1_50}
\end{figure*}

With an episode length of $H = 5$, it is easy for small differences in actions to yield noticeable differences in cumulative reward, as noted previously in the oil problem with Laplace survey function. Therefore, two agents of each algorithm were trained in an oil experiment with Laplace survey function with $\lambda = 1$, but with $H = 50$ time steps. This means that the maximum cumulative reward scales from $4.25$ to $49.25$, and that the SPAQL policy will be greedy for more time steps, allowing for existing balls to split further. The result is shown in Figure~\ref{fig:oil_laplace_1_50}. Both SPAQL agents stabilize at a maximum before $2000$ training iterations. The AQL agents, on the other hand, have a much fuzzier training curve, take longer to reach the best rewards, and use almost ten times more arms. Furthermore, looking at the partition for the first time step, it is clear that the action therein taken is not the optimal one. Even so, it should also be noticed that, given enough training iterations, the information about the global optimum (which is already available at time step $25$) would eventually propagate all way until time step $1$, thus yielding the optimal policy. This is a clear advantage of AQL over SPAQL. Given a time-invariant problem, if one of the time steps finds the optimal actions, the update rule will eventually propagate this information to all other time steps. However, this comes at a high cost in terms of memory (number of arms) and time (training iterations).

It is important to highlight that, in the first $1000$ training iterations, the cumulative reward for the AQL agents decreases steadily, while the number of arms is increasing. Even if given enough training iterations to allow the cumulative rewards to reach the maximum level, the arms associated with suboptimal policies will remain inside the partition, wasting memory and computational resources. This shows that, even for medium-sized episodes, AQL agents may not be a good choice for time-invariant problems, as growth in the number of arms is not necessarily followed by an increase in cumulative reward, and suboptimal arms remain in the policy even after a better one has been found

\subsection{Discussion}

In almost all experiments,  the SPAQL agents are less sensitive to the scaling parameter $\xi$ (the exception being the oil problem with Laplace survey function). Recalling that the scaling parameter controls the value of the bonus term, which is introduced to deal with the exploitation/exploration trade-off, this lower sensitivity to the scaling parameter is probably due to SPAQL using Boltzmann exploration to deal with the same trade-off. The lower sensitivity to this parameter, along with the fact that most environments shown tend to favor lower scaling values, suggest that the scaling term could probably be removed from SPAQL (by setting it to $0$). However, this would reduce SPAQL to Q-learning with Boltzmann exploration, which has not been proved to be sample-efficient. Therefore, the final recommendation of this paper with regards to the choice of the scaling value $\xi$ is to choose a non-zero value lower than $H/3$.

As a final note on the scaling parameter, recall Equation~\ref{eqn:conf_radius} for the expression of $\xi$ in terms of the other training variables. Neglecting the term proportional to the Lipschitz constant $L$, setting $H=5$, $K=1000$, and considering a low value of $\pfail$ (for example $0.1$), the value obtained for the scaling parameter is larger than $80$. This goes against the experimental evidence presented, which points towards smaller values of the scaling parameter being preferable over large ones, even in the AQL agents. It also exceeds $H$, which is the maximum value that $\Qhat{}{}(x,a)$ is allowed to have. A bonus term of $80$ in an environment where the episode length is $5$ would have no effect other than saturating the updates to the value function. In future work, it would be interesting to understand why this happens, and if the expression of the bonus term can be modified in order for it to be bounded by $H$.

\section{Conclusions}
\label{section:conclusion}
This paper introduces SPAQL, an improved version of AQL tailored for learning time-invariant policies. In order to balance exploration and exploitation, SPAQL uses Boltzmann exploration with a cyclic temperature schedule in addition to upper confidence bounds (UCB). Experiments show that, with very little parameter tuning, SPAQL performs satisfactorily in the problems where AQL was originally tested, resulting in partitions with a lower number of arms, and requiring fewer training iterations to converge. The two problems studied can be seen as a deterministic and a stochastic variant of the same problem, with SPAQL performing better than AQL on all but one of the instances of the stochastic problem.

There are several possible directions for further work. Both AQL and SPAQL can be further modified in several ways. The next natural modification would be to run several episodes during each training iteration, instead of only one. In SPAQL, this would increase the exploitatory behavior of the algorithm, possibly solving the cumulative reward difference seen in the oil problem with Laplace survey function. However, this would introduce another parameter (the number of episodes to run during each training iteration) into the algorithm, thus increasing the effort required for tuning. Currently, the parameters of SPAQL are highly conjugate with the episode length. It would also be interesting to study automatic ways of setting $u$ and $d$ given $H$. Similarly to what was done in this paper for $\xi$, an empirical study regarding the effect of tuning parameters $u$ and $d$ on SPAQL performance could be done.

Another parameter that may have a relevant impact is the number of evaluation rollouts, $N$. If this number is too low, the performance estimates may be inaccurate, and lead SPAQL to bad decisions when updating the stored partition. However, a larger $N$ also means more training time. On the other hand, as the state-action space is partitioned and the radius of the balls decreases, lower values of $N$ are required to obtain accurate estimates. Therefore, it would also be interesting to see if it is possible to set $N$ automatically based on the rollouts themselves.

Finally, a formal analysis of the complexity of SPAQL would allow a comparison in terms of sample efficiency to AQL or similar algorithms. On the experimental side, it would also be interesting to test the performance of SPAQL (along with AQL) in realistic settings like control applications, that typically have larger state and action spaces, and require a higher number of steps per episode.

    \section*{Acknowledgments}
    	We would like to thank \citet{Sinclair19} for making the code of AQL freely available, and for allowing us to use the LaTeX template of their paper as a template for this one. We would also like to thank Sean Sinclair for many conversations.
    	
    	This research work is supported by Fundação para a Ciência e Tecnologia, through IDMEC, under LAETA, project UIDB/50022/2020.
    \newpage
    \bibliographystyle{apalike}
    {\bibliography{spaql_bib}}

\newpage
\appendix

\renewcommand\thefigure{\thesection.\arabic{figure}}
\setcounter{figure}{0}

\section{Experimental results and figures}
\label{section:full_experiments}

In this section, the learning curves for the best agents found during the scaling experiments are shown. Each curve shows the average cumulative reward and the $95\%$ confidence interval. The scaling values are identified in the respective legend. The average number of arms is also shown, along with the state-action space partition for the best SPAQL agent at the end of training. Finally, the AQL agent partition at time steps $1$, $3$ and $5$ is also shown.

Each section contains some brief comments to the images, which complement the analysis in Sections~\ref{oil_experimental_results} and~\ref{ambulance_experimental_results}.

\subsection{Oil problem with quadratic survey function}

\begin{figure*}[!th]
\centering
\includegraphics[width=0.8\columnwidth]{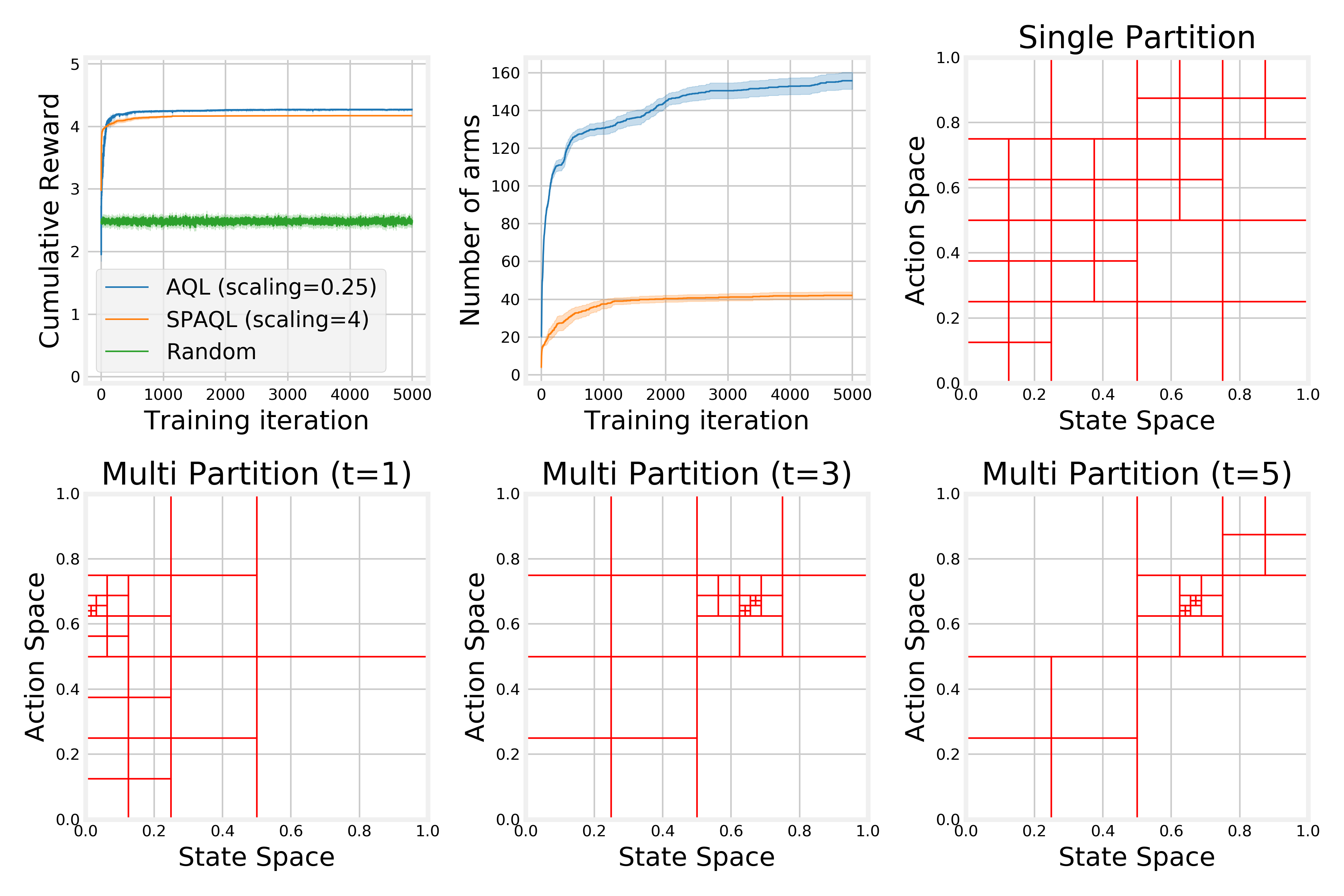}
\caption{
Comparison of the algorithms on the oil problem with quadratic survey function ($\lambda=1$). The SPAQL agents managed to reach a similar cumulative reward with one third of the arms.
}
\label{fig:oil_quadratic_1}
\end{figure*}

Starting with $\lambda=1$ (Figure~\ref{fig:oil_quadratic_1}), it is clear that the SPAQL agent did not exploit the optimal point of $0.75$, and over partitioned uninteresting regions of the space. The high rewards over a large area are probably a cause for this, since when the value of $\lambda$ is increased, concentrating the rewards, SPAQL agents partition the neighborhood of the optimal point more finely. Another possible cause is that the best performance currently stored happened by chance to be unexpectedly high. A way to prevent this from happening is to increase the number of evaluation rollouts $N$. In this experiment, $N = 20$ rollouts were used to evaluate the agents, but a higher number would lead to better estimates. The AQL agent learned that the optimal action given initial state $x=0$ is $a=0.75$, and then learned to hold the action $a=0.75$. The partitions in time instants $3$ and $5$ are very similar, as would be expected given that this is a time-invariant problem. It is this sort of partition duplication that is avoided by keeping only one partition.

\begin{figure*}[!th]
\centering
\includegraphics[width=0.8\columnwidth]{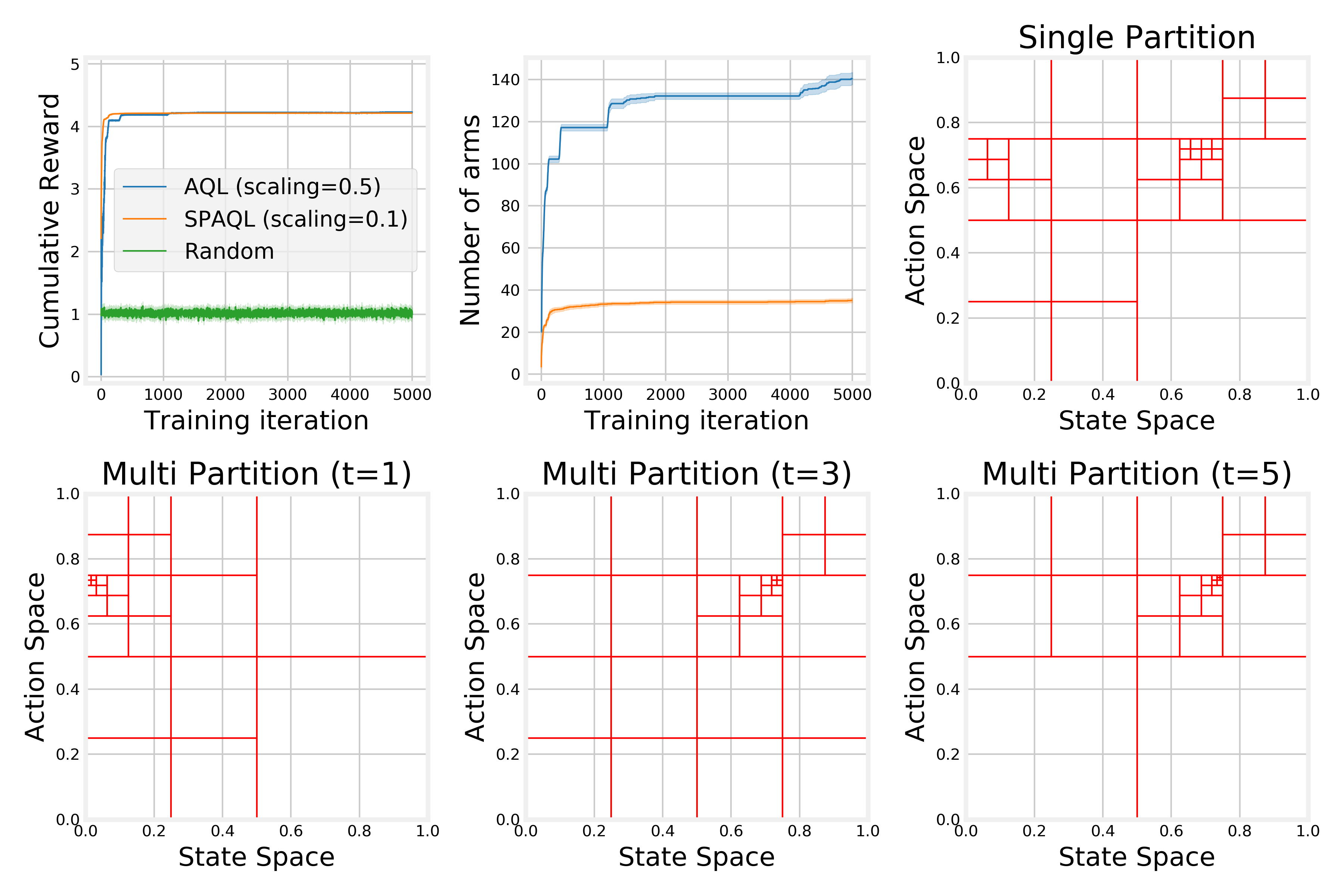}
\caption{
Comparison of the algorithms on the oil problem with quadratic survey function ($\lambda=10$).
}
\label{fig:oil_quadratic_10}
\end{figure*}

Moving to the case when $\lambda=10$ (Figure~\ref{fig:oil_quadratic_10}), the average cumulative rewards at the end of training for both algorithms are barely distinguishable. The SPAQL agents converge earlier than the AQL agents, and with fewer arms. The best SPAQL partition shows a better exploitation of the optimal point, compared with the case when $\lambda=1$. Once again very similar partitions are seen in time steps $3$ and $5$ in the best AQL agent. The increase in the number of AQL arms around training iteration $4100$ might seem to indicate that AQL agents have still not converged. However, the training curves indicate that both agents are very close to the maximum, meaning that the extra arms in the AQL agent correspond to new splits which are approximating the location of the optimal point to an accuracy of millimeters. The number of arms can grow indefinitely in order to approximate all of the decimal places in the floating point representation of $0.7+\pi/60$, and therefore should not be considered when assessing convergence of the algorithm.

\begin{figure*}[!th]
\centering
\includegraphics[width=0.8\columnwidth]{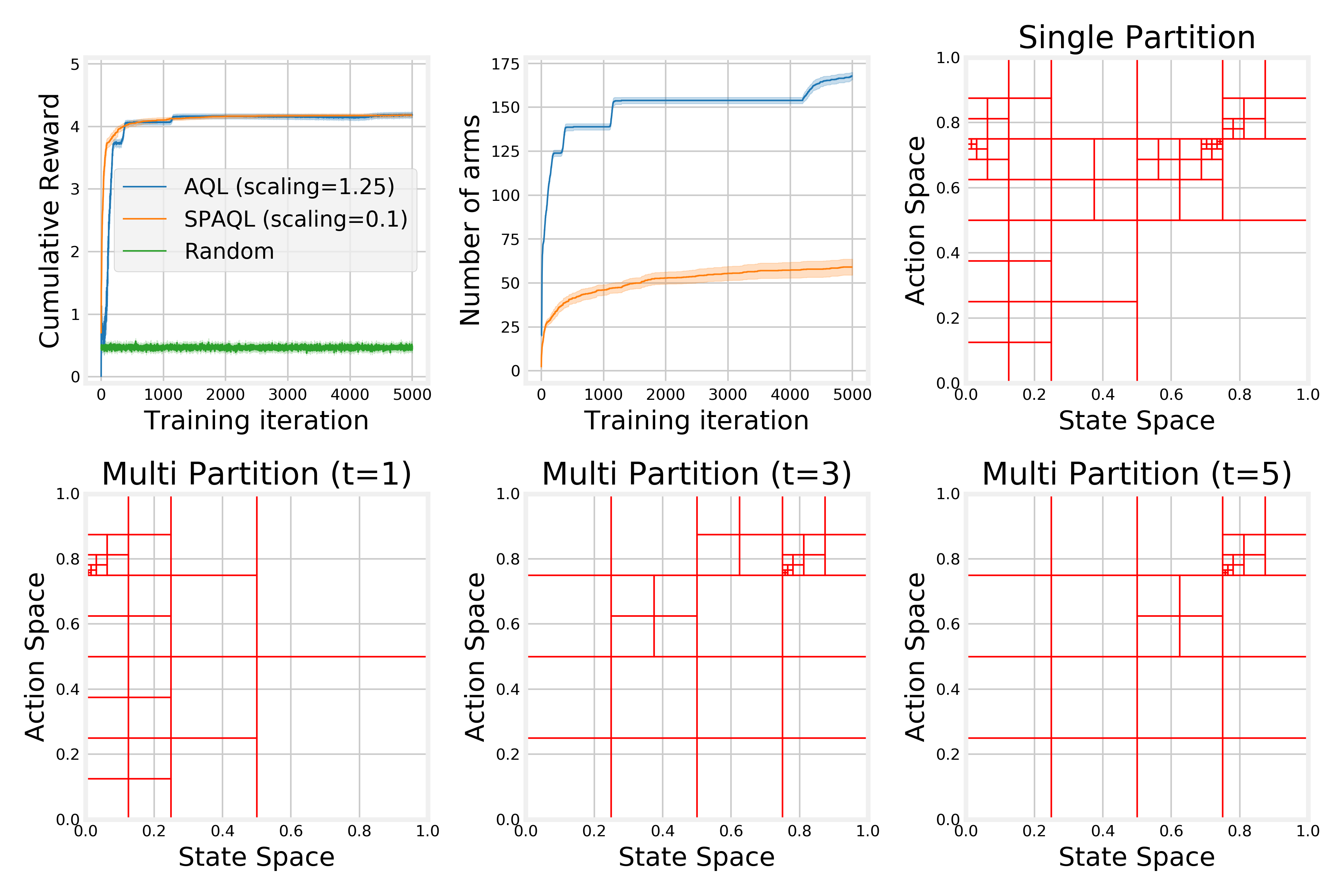}
\caption{
Comparison of the algorithms on the oil problem with quadratic survey function ($\lambda=50$).
}
\label{fig:oil_quadratic_50}
\end{figure*}

Finally, Figure~\ref{fig:oil_quadratic_50} shows the results when $\lambda=50$. These results are similar to the case of $\lambda=10$. The concentrated rewards have lead to an even better exploitation of the optimal point by the SPAQL agent.

\FloatBarrier
\clearpage
\subsection{Oil problem with Laplace survey function}

\begin{figure*}[!th]
\centering
\includegraphics[width=0.8\columnwidth]{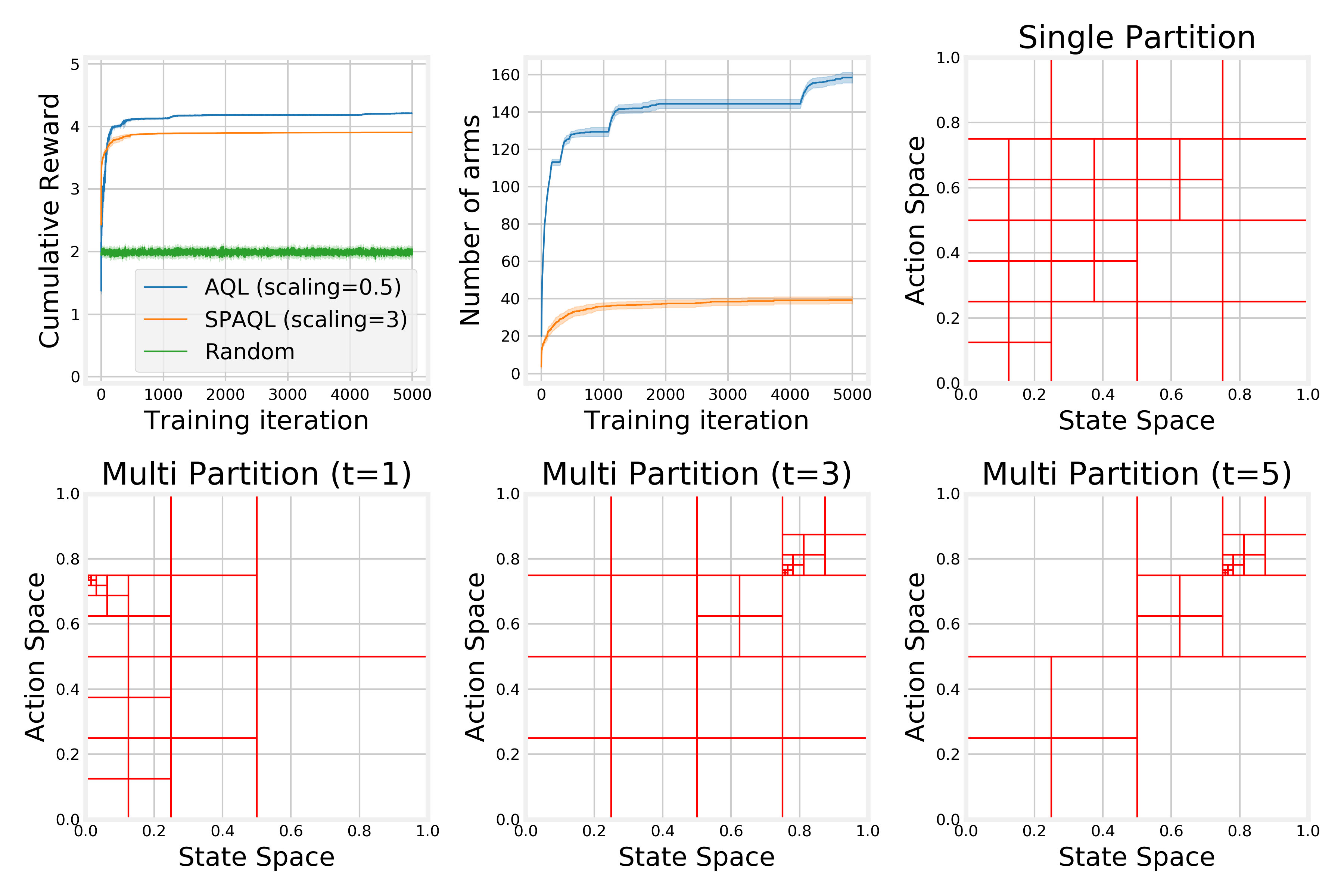}
\caption{
Comparison of the algorithms on the oil problem with Laplace survey function ($\lambda=1$).
}
\label{fig:oil_laplace_1}
\end{figure*}

When using the Laplace survey function rewards become more concentrated than when using the quadratic survey function. This is the main reason behind the differences in cumulative reward at the end of training. Considering the case with $\lambda=1$ (Figure~\ref{fig:oil_laplace_1}), although the SPAQL agents stabilize at slightly lower rewards, they still manage to achieve them with a much smaller number of arms. The partition for the best SPAQL agent shows a lack of exploitation of the optimal point, as was the case with the quadratic survey function with $\lambda=1$ (Figure~\ref{fig:oil_quadratic_1}). The partitions for time step $3$ and $5$ in the AQL agent are once again barely distinguishable.

\begin{figure*}[!th]
\centering
\includegraphics[width=0.8\columnwidth]{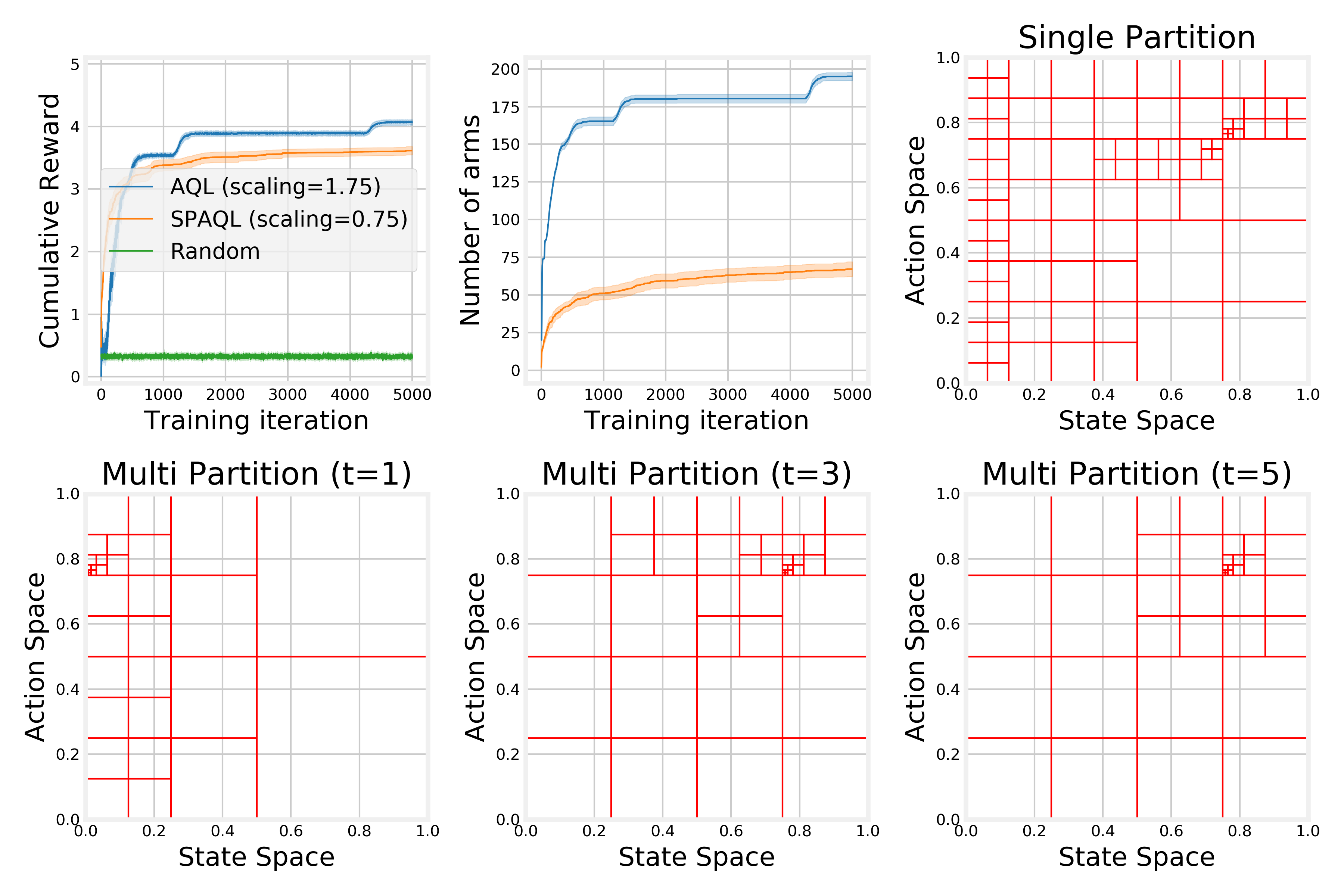}
\caption{
Comparison of the algorithms on the oil problem with Laplace survey function ($\lambda=10$).
}
\label{fig:oil_laplace_10}
\end{figure*}

Moving to $\lambda=10$ (Figure~\ref{fig:oil_laplace_10}), training has become more difficult for both types of agent, when compared with the case when $\lambda = 1$. The partition of the SPAQL agent indicates a lot of exploration from the initial state ($x=0$), along with a fine partition around the optimal point, meaning that, once found, it was exploited. Since the rewards are more concentrated, and the AQL agent partitions the neighborhood of the optimal point more finely, it ends up collecting higher rewards than the SPAQL agent, which keeps a coarser partition around the optimal point. However, it should be noted that, for all practical purposes, the SPAQL agent managed to find the approximate location of the optimal point, using fewer arms than the AQL agent.

\begin{figure*}[!th]
\centering
\includegraphics[width=0.8\columnwidth]{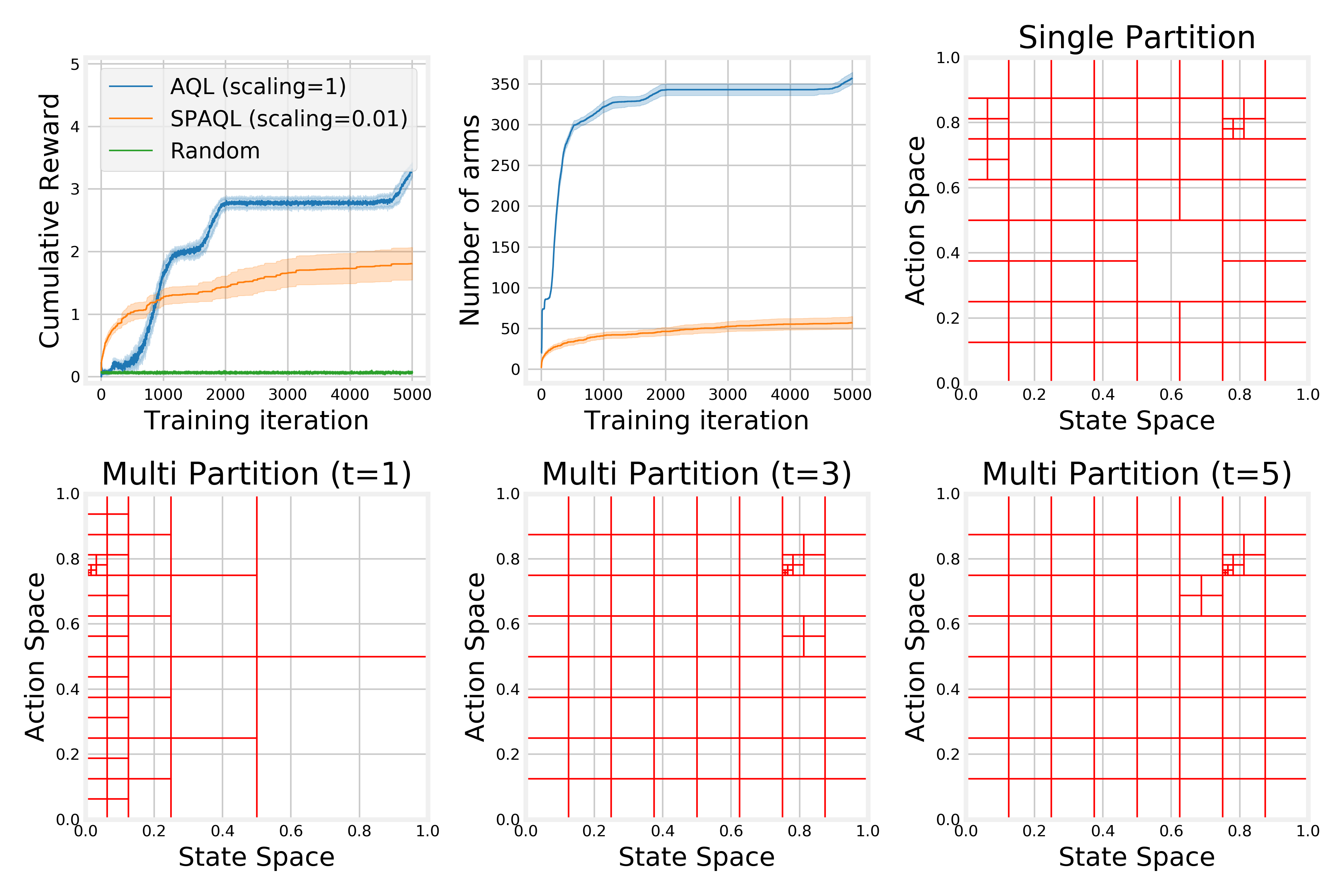}
\caption{
Comparison of the algorithms on the oil problem with Laplace survey function ($\lambda=50$).
}
\label{fig:oil_laplace_50}
\end{figure*}

Finally, Figure~\ref{fig:oil_laplace_50} shows the results when $\lambda=50$. This is the hardest problem, with the most concentrated rewards. Both algorithms locate the optimal point, but the AQL agents exploit it much more, leading to higher rewards. However, in the process, it partitioned a lot the individual partitions, leading to an average number of arms six times higher than the ones in the SPAQL partition, which has implications regarding the amount of resources required to store all the partitions of the agent.

In the two latter cases ($\lambda \in \{10, 50\}$), the $5000$ iterations are clearly not enough for the AQL agents to converge. However, if finding the approximate location of the oil deposit is considered as the objective of this problem, then the $5000$ iterations were enough. The cumulative rewards are lower when compared to other episodes due to the high concentration of rewards around the optimal point, and the lack of reward signal in the remaining state-action space.

\FloatBarrier
\clearpage
\subsection{Ambulance problem with uniform arrivals}

\begin{figure*}[!th]
\centering
\includegraphics[width=0.8\columnwidth]{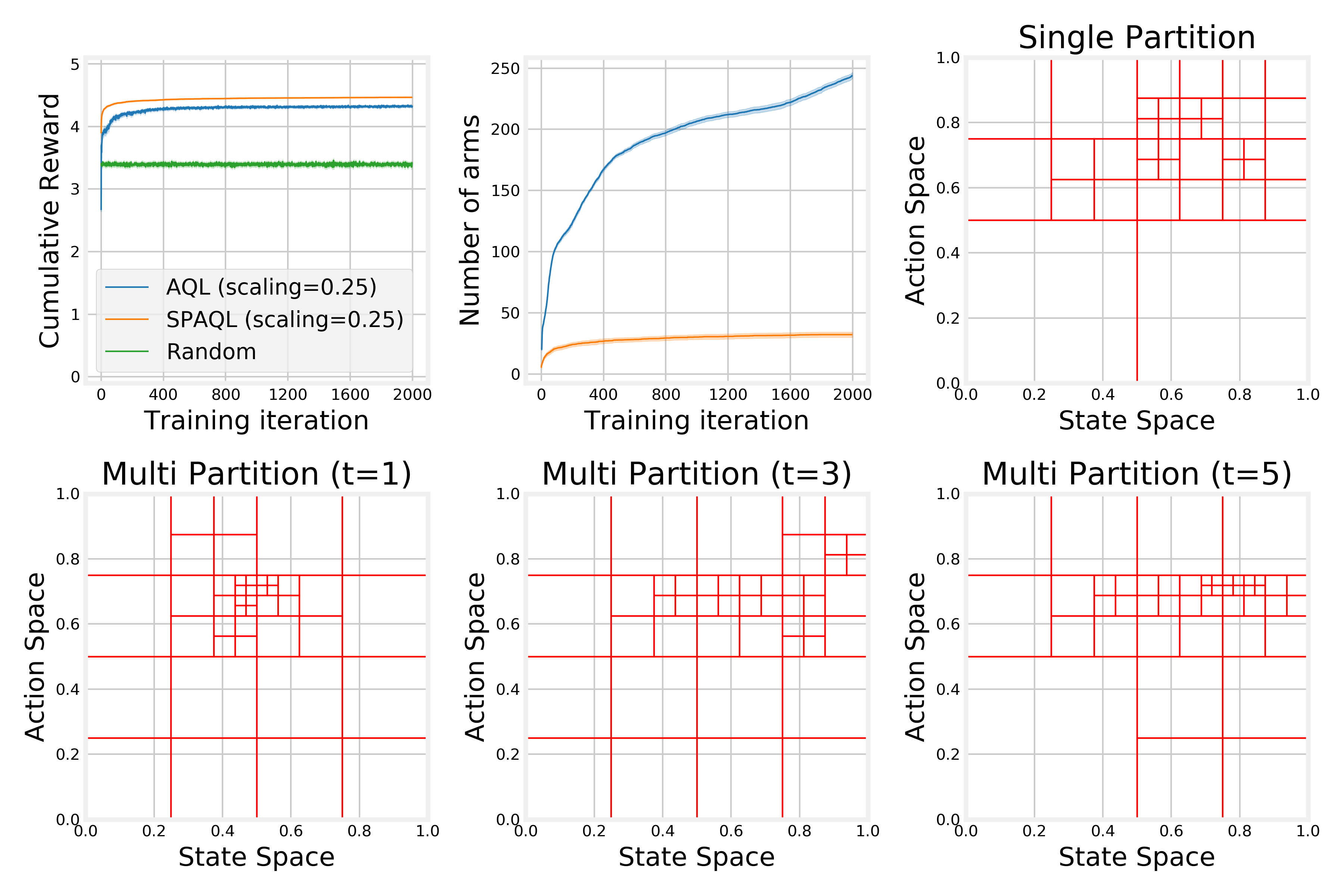}
\caption{
Comparison of the algorithms on the ambulance problem with uniform arrival distribution and only paying the cost to relocate ($c=0$).
}
\label{fig:ambulance_uniform_0}
\end{figure*}

For the case when $c=0$, shown in Figure~\ref{fig:ambulance_uniform_0}, the heuristic solution to this problem is the ``Mean'' heuristic, which corresponds to the line $a=0.5$. The SPAQL agents managed to achieve higher rewards with a coarser partition and about one fifth of the arms of the AQL agents.

\begin{figure*}[!th]
\centering
\includegraphics[width=0.8\columnwidth]{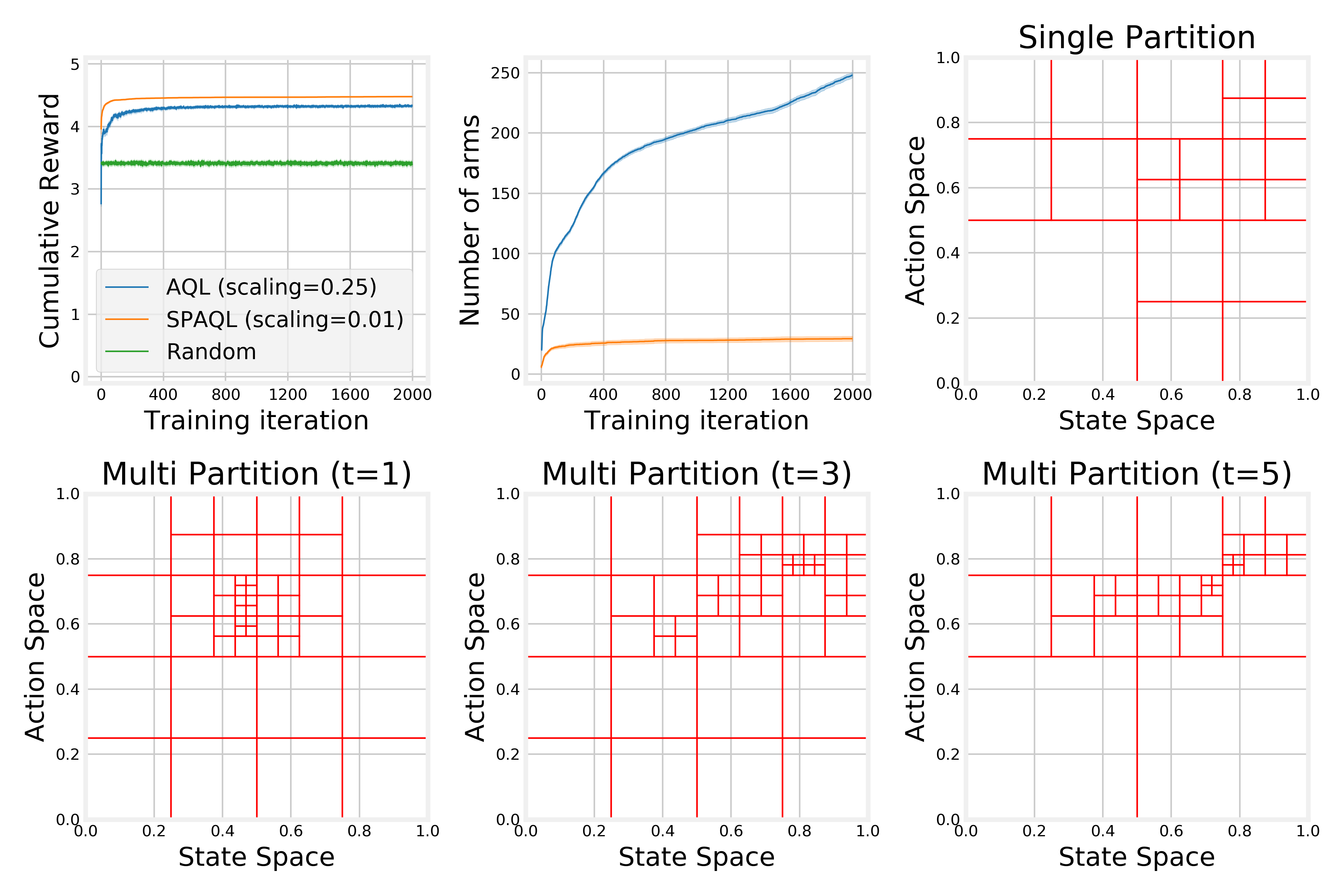}
\caption{
Comparison of the algorithms on the ambulance problem with uniform arrival distribution and paying a mix between the cost to relocate and the cost to go ($c=0.25$).
}
\label{fig:ambulance_uniform_0.25}
\end{figure*}

Moving to Figure~\ref{fig:ambulance_uniform_0.25}, corresponding to $c=0.25$, the optimal policy would be a mix of the ``Mean'' (horizontal line) and the ``No Movement'' (diagonal line) heuristic. The SPAQL agents managed to achieve higher rewards with a coarser partition and less than one fifth of the arms of the AQL agents, a reduction even greater than the one in the case $c=0$.

\begin{figure*}[!th]
\centering
\includegraphics[width=0.8\columnwidth]{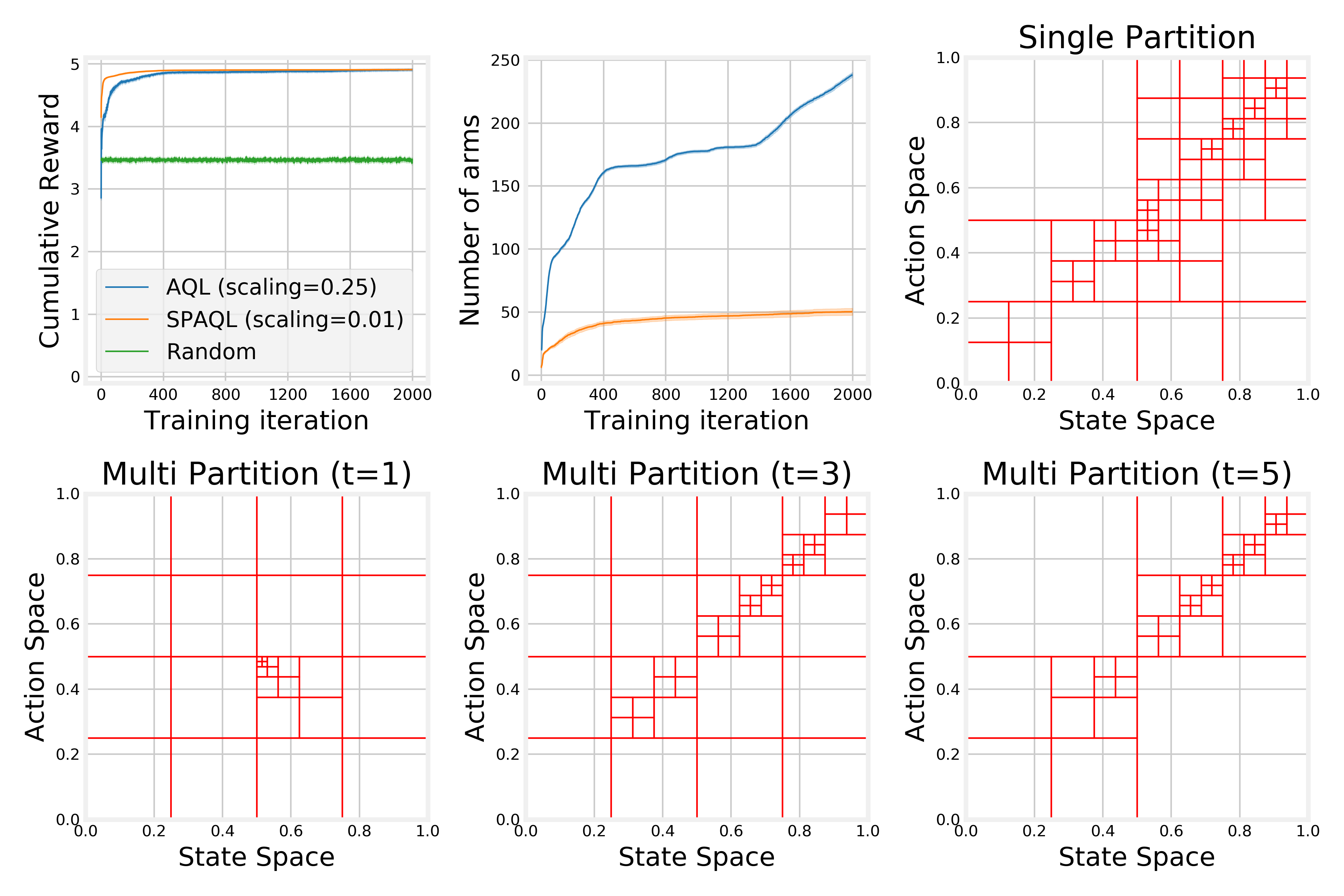}
\caption{
Comparison of the algorithms on the ambulance problem with uniform arrival distribution and paying only the cost to go ($c=1$).
}
\label{fig:ambulance_uniform_1}
\end{figure*}

Finally, Figure~\ref{fig:ambulance_uniform_1} shows the results when $c=1$. The optimal policy would be the ``No Movement'' (diagonal line) heuristic. The diagonal line appears finely partitioned in both types of agents. However, there was clearly a shortage of samples of states within $[0, 0.25]$ in time steps $3$, and $5$, which may bring a problem in deployment. The SPAQL agents managed to achieve higher rewards with around one fifth of the arms of the AQL agents.

\FloatBarrier
\clearpage

\subsection{Ambulance problem with beta arrivals}

\begin{figure*}[!th]
\centering
\includegraphics[width=0.8\columnwidth]{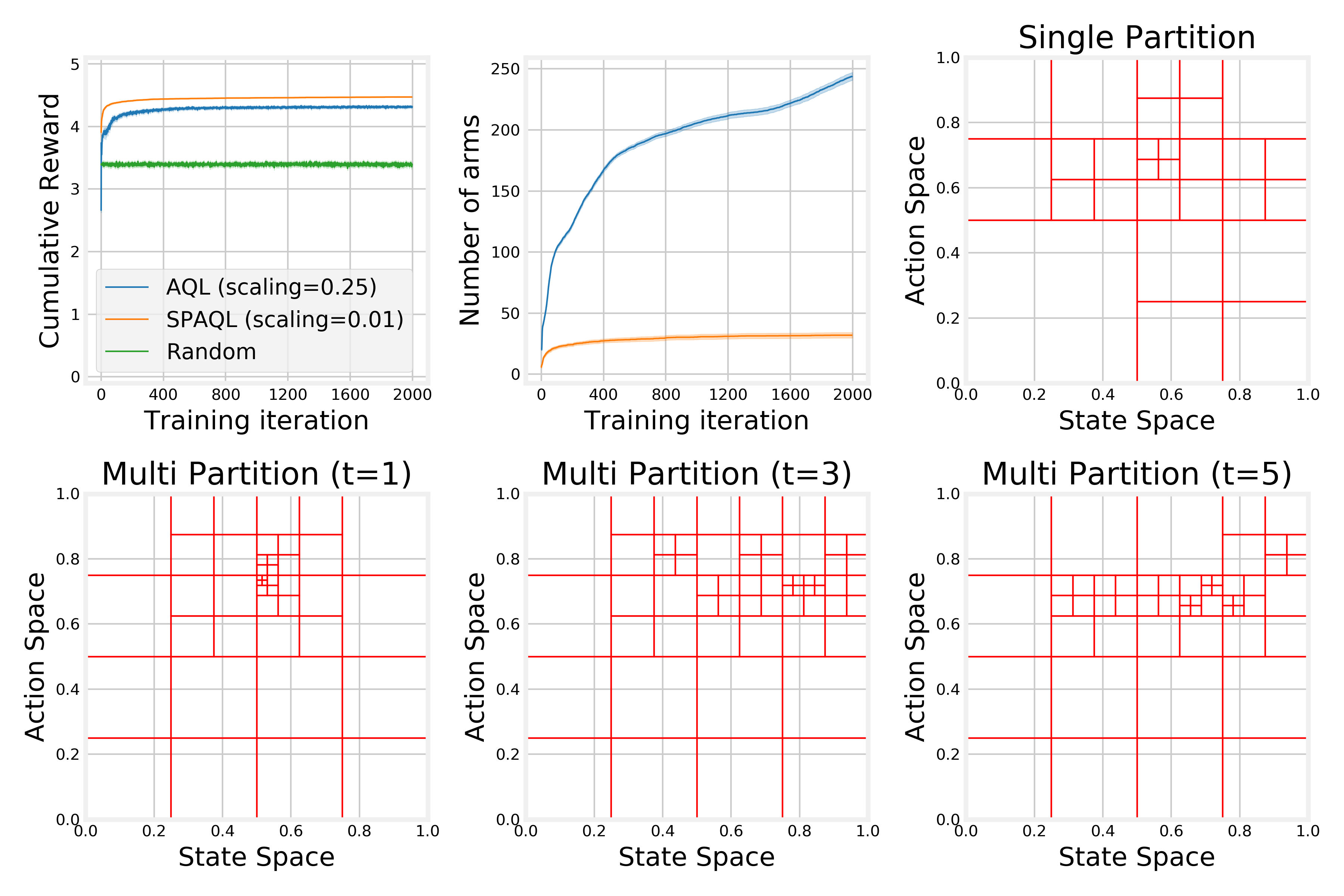}
\caption{
Comparison of the algorithms on the ambulance problem with $\text{Beta}(5,2)$ arrival distribution and only paying the cost to relocate ($c=0$).
}
\label{fig:ambulance_beta_0}
\end{figure*}

For the case when $c=0$, shown in Figure~\ref{fig:ambulance_beta_0}, the heuristic solution to this problem is the ``Mean'' heuristic, which corresponds to the line $a\approx0.7$. The SPAQL agent managed to achieve higher rewards with a coarser partition and less than one fifth of the arms of the AQL agents.

\begin{figure*}[!th]
\centering
\includegraphics[width=0.8\columnwidth]{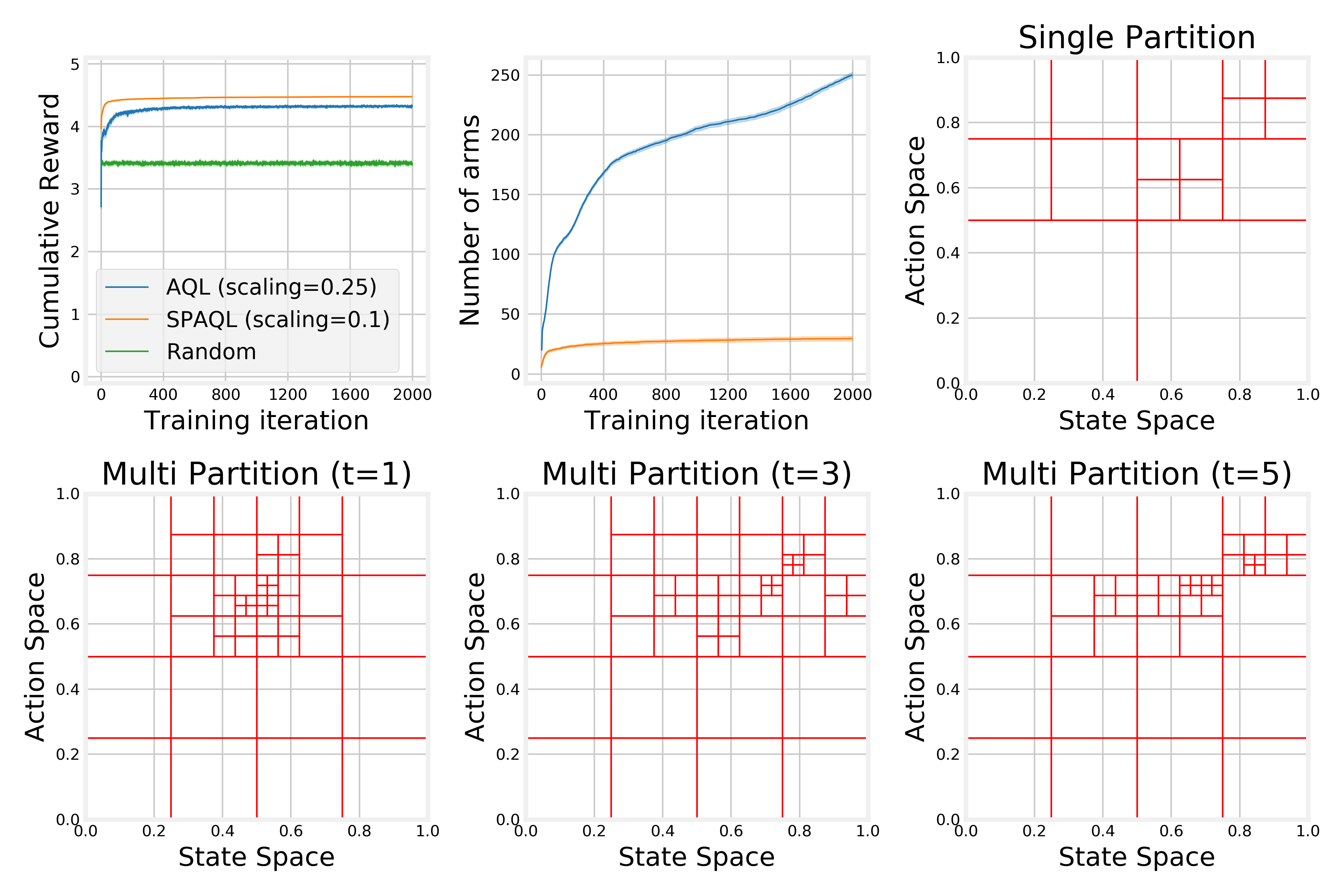}
\caption{
Comparison of the algorithms on the ambulance problem with $\text{Beta}(5,2)$ arrival distribution and paying a mix between the cost to relocate and the cost to go ($c=0.25$).
}
\label{fig:ambulance_beta_0.25}
\end{figure*}

Moving to Figure~\ref{fig:ambulance_beta_0.25}, corresponding to $c=0.25$, the optimal policy would be a mix of the ``Mean'' (horizontal line) and the ``No Movement'' (diagonal line) heuristic. The SPAQL agent managed to achieve higher rewards with a coarser partition and less than one fifth of the arms of the AQL agents.

\begin{figure*}[!th]
\centering
\includegraphics[width=0.8\columnwidth]{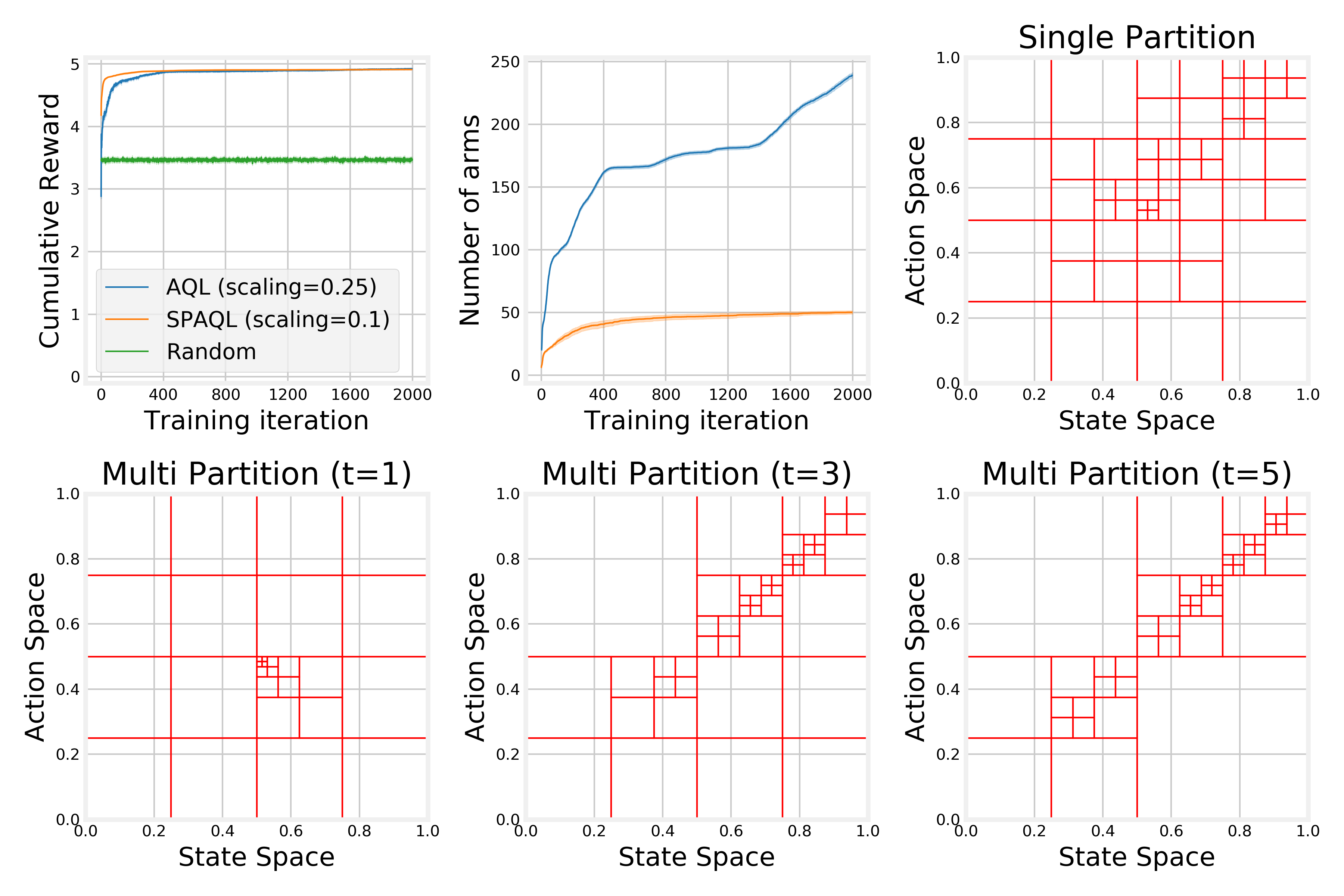}
\caption{
Comparison of the algorithms on the ambulance problem with $\text{Beta}(5,2)$ arrival distribution and paying only the cost to go ($c=1$).
}
\label{fig:ambulance_beta_1}
\end{figure*}

Finally, Figure~\ref{fig:ambulance_beta_1} shows the results when $c=1$. The optimal policy would be the ``No Movement'' (diagonal line) heuristic. Since arrivals are now concentrated around $0.7$ (the mean of the distribution), the diagonal appears more partitioned for states within $[0.5, 1]$ than for states within $[0, 0.5]$. The AQL agents do a finer partition of the diagonal, but this still does not prevent the SPAQL agents from converging to better rewards earlier, and with around one fifth of the number of arms of the AQL agents.

\FloatBarrier
	
\end{document}